\documentclass{article}
\newif\ifpreprint\preprinttrue
\newif\ifpagecut\pagecutfalse
\usepackage{iclr2027_conference,times}
\usepackage[utf8]{inputenc}
\usepackage{graphicx}
\usepackage{amsmath,amssymb}
\usepackage[colorlinks=true, allcolors=blue]{hyperref} % link colours as in the PNAS class
\usepackage{url}
\DeclareUnicodeCharacter{2192}{\ensuremath{\rightarrow}} % "A→E tie-break" in the human-study text

\newcommand{\PaperTitle}{Stance Drift: How AI-Mediated Communication Distorts Our Message}

\makeatletter
\newlength{\prompt@xshift}      
\newlength{\prompt@toppad}      
\newlength{\prompt@rulethick}   
\newlength{\prompt@titlegap}    
\newlength{\prompt@rulebelowsep}

\newlength{\prompt@titleboxrule}
\newlength{\prompt@titleboxsep} 
\newcommand{\prompttitleformat}[1]{%
  \begingroup
    \fboxsep=\prompt@titleboxsep \fboxrule=\prompt@titleboxrule
    \fbox{\bfseries\textsc{#1}}%
  \endgroup
}

\newsavebox{\prompt@titlebox}
\newlength{\prompt@titlesink}
\newlength{\prompt@ruleoffset}

\newenvironment{promptlines}[1][]%
{%
  \par
  \setlength{\prompt@titlesink}{0pt}%
  \setlength{\prompt@ruleoffset}{0pt}%
  \def\prompt@has@title{0}%
  \ifx\relax#1\relax\else
    \sbox{\prompt@titlebox}{\prompttitleformat{#1}}%
    \setlength{\prompt@titlesink}{.5\ht\prompt@titlebox}%
    \setlength{\prompt@ruleoffset}{%
      \dimexpr \prompt@xshift+\wd\prompt@titlebox+\prompt@titlegap\relax}%
    \def\prompt@has@title{1}%
  \fi
  \noindent
  \ifnum\prompt@has@title=1
    \makebox[0pt][l]{%
      \hspace{\prompt@xshift}%
      \raisebox{-\prompt@titlesink}{\usebox{\prompt@titlebox}}%
    }%
  \fi
  \hspace*{\prompt@ruleoffset}%
  \leaders\hrule height\prompt@rulethick\hfill\kern0pt\par
  \vspace*{\dimexpr \prompt@toppad+\prompt@titlesink+\prompt@rulebelowsep\relax}%
  \begingroup
    \ttfamily
    \parindent=0pt \parskip=0pt
    \raggedright\sloppy
    \hyphenchar\font=`\-
    \emergencystretch=3em
    \rightskip=0pt plus 2em
    \obeylines
}%
{%
  \par\endgroup
  \vspace*{\prompt@rulebelowsep}%
  \noindent\leaders\hrule height\prompt@rulethick\hfill\kern0pt\par
}
\makeatother

\graphicspath{{./}}
\newcommand{\SPRGPTfouromini}{0.570}

\newcommand{\SPRGPTfourominiReversed}{0.584}

\newcommand{\SPRGPTfourominiTempZero}{0.571}

\newcommand{\SPRGPTfourone}{0.686}

\newcommand{\SPRGPTfivefour}{0.593}

\newcommand{\SuccessRateAgreeGPTfivefour}{0.147}

\newcommand{\SuccessRateDisagreeGPTfivefour}{0.136}

\newcommand{\FlipMassAgreeGPTfivefour}{0.076}

\newcommand{\PolMassGPTfivefour}{0.328}

\newcommand{\SuccessRateDisagreeStronglyLlamaFourMaverick}{0.515}

\newcommand{\SPRGPTfourominiReflection}{0.586}

\newcommand{\SPRGPTfivefourReflection}{0.613}

\newcommand{\SPRGPTfivefourThirdPersonReflection}{0.637}

\newcommand{\SuccessRateNeutralGPTfivefourThirdPersonReflection}{0.721}

\newcommand{\FlipMassAgreeGPTfivefourThirdPersonReflection}{0.252}

\newcommand{\SPRGPTfivefourThirdPersonReflectionMedium}{0.775}

\newcommand{\SuccessRateNeutralGPTfivefourThirdPersonReflectionMedium}{0.607}

\newcommand{\FlipMassAgreeGPTfivefourThirdPersonReflectionMedium}{0.010}

\newcommand{\PolMassGPTfivefourThirdPersonReflectionMedium}{0.119}

\newcommand{\SPRGPTfourominiAssert}{0.559}

\newcommand{\SPRGPTfourominiMultipleExtraction}{0.569}

\newcommand{\SPRGPTfourominiInContext}{0.573}

\newcommand{\FailedEncodeCellsWord}{six}
\newcommand{\HumanExtractionAgreement}{0.743}
\newcommand{\HumanExtractionAgreementSE}{0.038}
\newcommand{\AIExtractionAgreement}{0.472}
\newcommand{\AIExtractionAgreementSE}{0.002}
\newcommand{\HumanExtractionChanceFloor}{0.222}
\newcommand{\DriftExampleTiesFlipWord}{four}
\newcommand{\DriftExampleTiesDevWord}{three}
\newcommand{\BonferroniBelowCells}{37}

\urldef{\CodeURL}\url{https://github.com/Lilacllc/LLM_stance_drift_debate_speeches.git}
\newcommand{\SurveyAuthor}{L.L.}

\title{\PaperTitle}

\newcommand{\AuthorBlock}[1]{\begin{tabular}[t]{@{}l@{}}#1\end{tabular}}
\author{\normalfont
\begin{tabular}[t]{@{\hspace{-3.5pt}}l@{\hspace{10pt}}l@{\hspace{-3.5pt}}}
\AuthorBlock{\bfseries Lingchong Liu \\
Department of Information Systems, \\
Business Statistics and Operations Management \\
Hong Kong University of Science and Technology \\
Hong Kong SAR} &
\AuthorBlock{\bfseries Yanfei Zhou \\
Department of Data Sciences and Operations \\
University of Southern California \\
Los Angeles, CA, USA} \\ \noalign{\vskip 12pt}
\AuthorBlock{\bfseries Jacob Bien \\
Department of Data Sciences and Operations \\
University of Southern California \\
Los Angeles, CA, USA} &
\AuthorBlock{\bfseries Y.X. Rachel Wang \\
School of Mathematics and Statistics \\
University of Sydney \\
Sydney, Australia} \\ \noalign{\vskip 12pt}
\AuthorBlock{\bfseries Lucy Xia \\
Department of Information Systems, \\
Business Statistics and Operations Management \\
Hong Kong University of Science and Technology \\
Hong Kong SAR} &
\AuthorBlock{\bfseries Xin Tong\thanks{Corresponding author. E-mail: \texttt{xint@hku.hk}} \\
Department of Data Sciences and Operations \\
University of Southern California \\
Los Angeles, CA, USA \\
Faculty of Business and Economics \\
University of Hong Kong, Hong Kong SAR}
\end{tabular}}

\iclrfinalcopy % preprint: authors shown, no line numbers; the header is set to "Preprint" below

\begin{document}

\maketitle
\lhead{Preprint}

\begin{abstract}
% abstract.tex -- the abstract body, input inside the abstract environment by both drivers.
Large language models (LLMs) increasingly mediate human communication, from drafting emails to summarizing scientific reports, yet whether they faithfully preserve a speaker's position remains largely untested.
We model AI-mediated communication as a two-step generation-extraction pipeline: one LLM produces an argument from a specified stance, and a second LLM extracts the stance from that argument.
We represent the pipeline as a probabilistic state transition over five Likert-type stance categories and define the stance preservation rate (SPR) as the average probability that the extracted stance matches the initial stance.
Across 112 debate propositions, none of the nine LLMs tested exceeded an SPR of $0.7$ under the default configuration. % lint:ok
Three drift patterns accounted for most of the drift: polarization, deviation from neutrality, and flipping.
Among the mitigation strategies tested, including in-context learning, multiple extraction with shuffled options, assertion, and reflection, only adding medium reasoning effort to a reflection prompt for \texttt{GPT-5.4} substantially improved the SPR, to $\SPRGPTfivefourThirdPersonReflectionMedium$, yet polarization remained the largest pattern, with $\PolMassGPTfivefourThirdPersonReflectionMedium$ of the transition mass.
An exploratory comparison with human extraction on a single proposition suggests that drift arises at both the generation and the extraction stage.
These results point to a fidelity gap in AI-mediated communication, with implications for journalism, policy deliberation, scientific communication, and other domains where opinion-laden messages pass through language models.

\end{abstract}

\section{Introduction}\label{sec:introduction}

\begin{figure}[htp]
    \centering
    \setlength{\tabcolsep}{2pt}% the three panels sum to 0.9\textwidth; the default 6pt padding wrapped panel C to a second row (TASK 4.2)
    \begin{tabular}[b]{c}
        \includegraphics[width=.3373\textwidth,trim=0.3cm 0cm 0.3cm 0cm,clip]{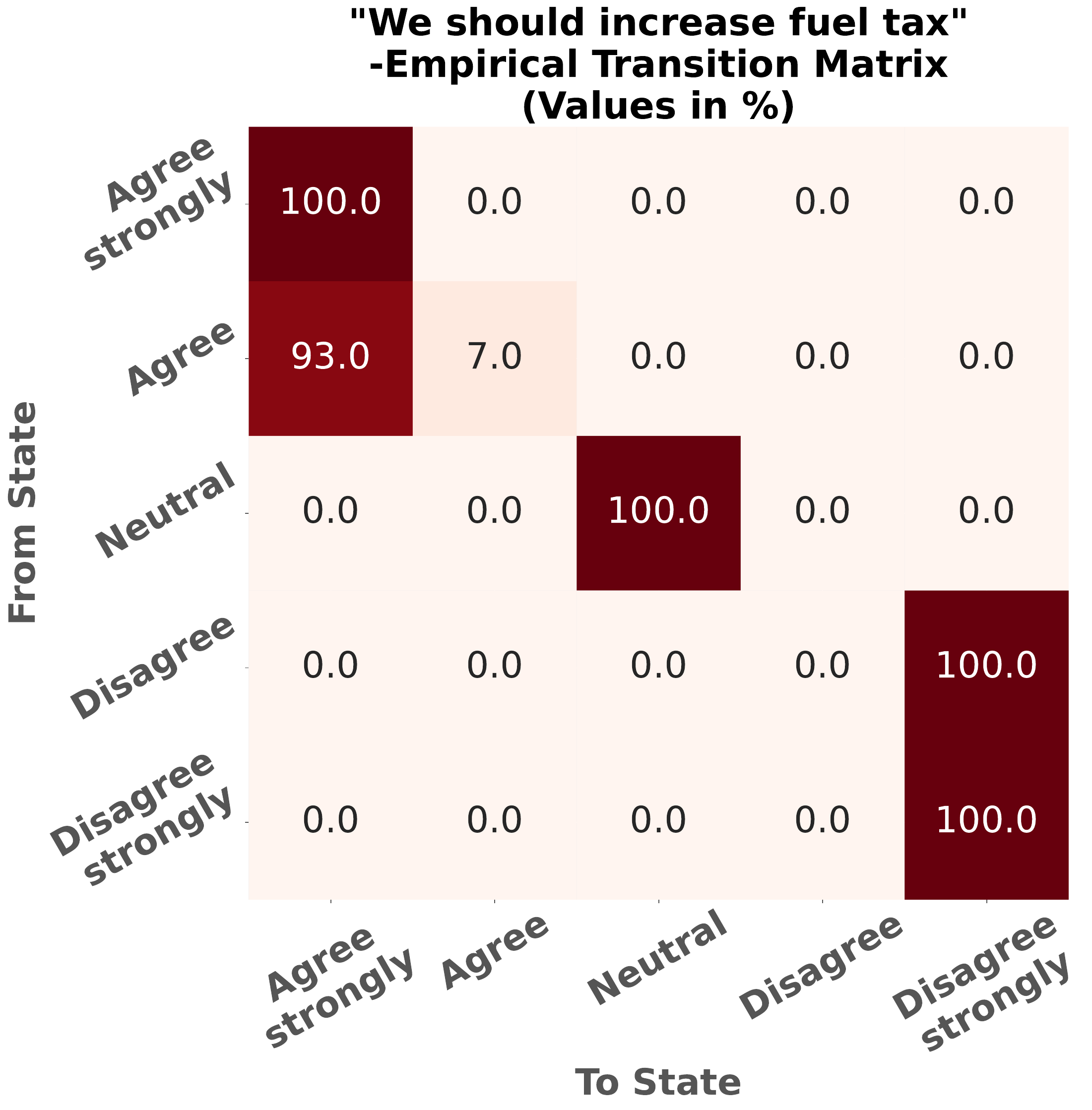} \\ 
        {\small (\textbf{A})}
    \end{tabular}%
    \begin{tabular}[b]{c}
      \includegraphics[width=.2554\textwidth,trim=0.3cm 0cm 0.3cm 0cm,clip]{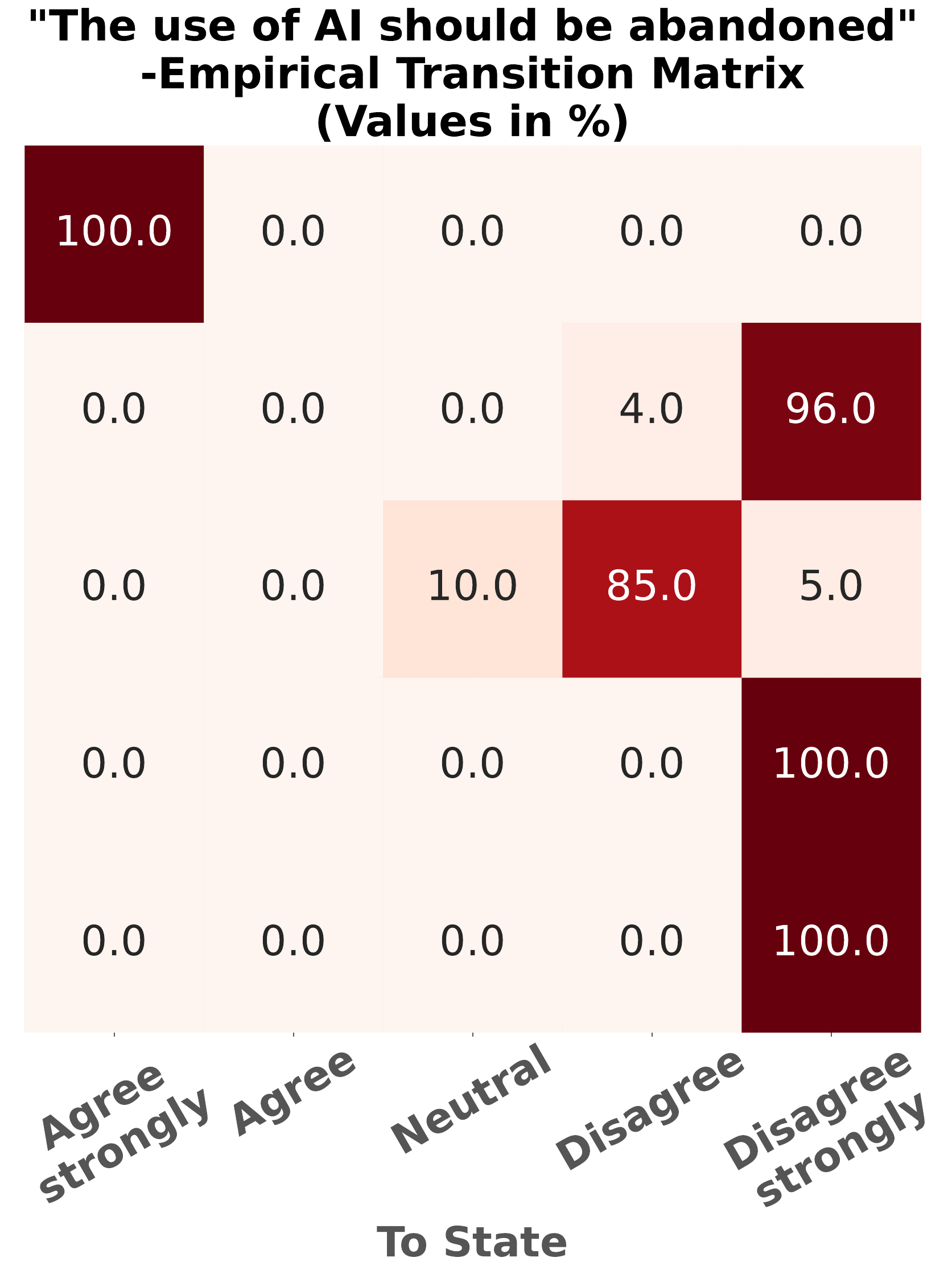} \\ 
      {\small (\textbf{B})}
    \end{tabular}%
    \begin{tabular}[b]{c}
      \includegraphics[width=.3073\textwidth,trim=0.3cm 0cm 0.3cm 0cm,clip]{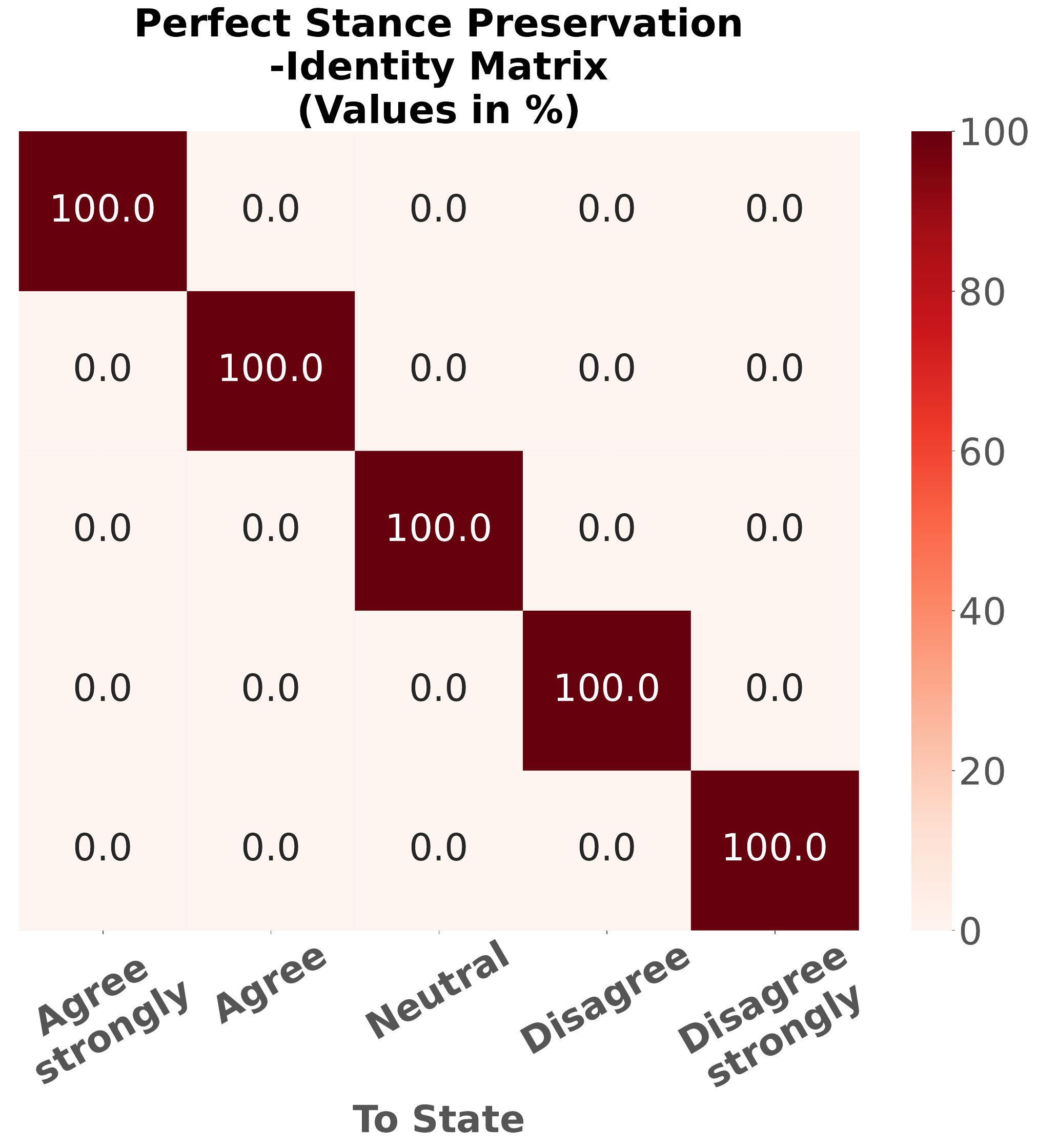} \\
      {\small (\textbf{C})}
    \end{tabular}
    \caption{
        \textbf{Example of empirical transition matrices deviating from the ideal identity matrix.}
        Rows are the initial stances and columns the extracted stances; the $(i,j)$ entry is the probability, in percent, of extracting the $j$th choice given the $i$th choice as the initial stance.
        (\textbf{A}) \texttt{GPT-5.4}, default configuration, proposition ``We should increase fuel tax''.
        (\textbf{B}) \texttt{GPT-5.4}, default configuration, proposition ``The use of AI should be abandoned''.
        (\textbf{C}) The identity matrix: perfect preservation of every stance.
    }
    \label{fig:teaser}
\end{figure}

History shows that small failures in message fidelity can have outsized consequences: the Berlin Wall fell hours after a confused spokesperson mistakenly announced that the borders were opening `immediately, without delay'~\citep{hertle2001fall}. Today, this fragility of transmission is no longer limited to human intermediaries but is now embedded in the algorithms that are increasingly integrated into human communication across domains including decision making~\citep{chen2023emergence, brand2024using}, scientific research~\citep{horton2023large, qin2024aiturk}, and communications~\citep{bick2024rapid, chatterji2025people}. In one illustrative example from our framework, an AI-mediated pipeline can invert an initial ``Agree'' stance on the proposition ``The use of AI should be abandoned'' into ``Disagree'' and ``Disagree strongly'', effectively reversing the speaker's position (Fig.~\ref{fig:teaser}B). The inversion evokes familiar science-fiction tropes about AI resisting shutdown, but the scientific issue here is more basic: whether large language models (LLMs) preserve a communicator's intent as messages pass through them.

This question is increasingly important because LLMs are now routine intermediaries in written communication. From July 2024 to June 2025, writing-related activities (including text generation, editing, critiques, summarization, and translation) made up more than 40\% of ChatGPT's work-related use~\citep{chatterji2025people}. Beyond the well-studied issues with LLM-generated content, such as inherent biases and hallucinations~\citep{acerbi2023large, spitale2023ai, piao2025emergence, rottger2025issuebench, wang2025echoes}, an equally important but less explored challenge is the preservation of \textit{stance}, the attitude or position toward a given proposition, during AI-mediated communication. % lint:ok

Classic stance detection frames this task as classifying a writer's position toward a target using discrete categories (e.g., in favor, against, or neutral), leading to benchmarks such as \texttt{SemEval-2016}~\citep{mohammad2016semeval} and surveys that consolidate conventions across platforms and domains~\citep{kucuk2020survey,aldayel2021stance}.
More recent work on subjective global opinions shows that LLM interactions often involve value-laden judgments, with model responses reflecting specific human opinion distributions~\citep{durmus2023towards,santurkar2023whose}, and examines LLM response behavior in surveys~\citep{tjuatja2024llms}, self-preference~\citep{panickssery2024llm}, and sycophancy~\citep{sharma2024towards}.
At the communication level, \citet{hancock2020ai} propose studying how AI intermediaries affect human communication, and related research studies whether transformations preserve perspective: co-writing with an opinionated model shifts writers' views~\citep{jakesch2023co}, and in news summarization, \citet{liu2024p3sum} find that summarization systems frequently alter an article's political stance.
Beyond single-step transformations, transmission-chain (``telephone game'') studies with LLMs reveal cumulative, directional changes as messages are iteratively regenerated~\citep{acerbi2023large,perez2024telephone}.
While valuable, these strands (i) do not capture the \textit{full} opinion-communication process from initial message generation to downstream stance extraction, (ii) focus on specific tasks or settings rather than fundamental mechanisms of communication, and (iii) often lack a general, quantitative statistic for end-to-end stance fidelity.

We present a systematic framework for quantifying \textit{stance} preservation in complete AI-mediated human-to-human communication processes by adopting a simple, task-agnostic decomposition—\emph{proposition} (what is asserted) plus \emph{stance} (position toward it)—and purposefully using discrete categories to match social-science measurement practices (Likert-type elicitation)~\citep{likert1932,croasmun2011using}. As LLM-generated text is repeatedly reused, real-world information flow naturally forms a generation-extraction pipeline (Fig.~\ref{fig:model}): a sender prompts a model to produce an argument from a given stance, and a receiver later extracts the stance from that text with LLM assistance. We therefore study whether the sender’s stance is faithfully conveyed through AI mediation. We call this \textbf{stance preservation} (the receiver recovers the same discrete stance as the sender intended) and refer to any AI-induced change as \textbf{stance drift}. We model the end-to-end process as a probabilistic state transition, introducing a transition matrix to quantify stance preservation: diagonal elements indicate the probability of retaining the original stance (with a value of $1$ representing perfect fidelity), while off-diagonal entries capture the extent of \textit{stance drift}. 

\ifpagecut\else
% fig_model.tex -- the AI-mediated communication model figure (main text of the preprint; appendix of the ICLR build, TASK 4.2).
\begin{figure}[t]
    \centering
    \includegraphics[width=0.9\textwidth]{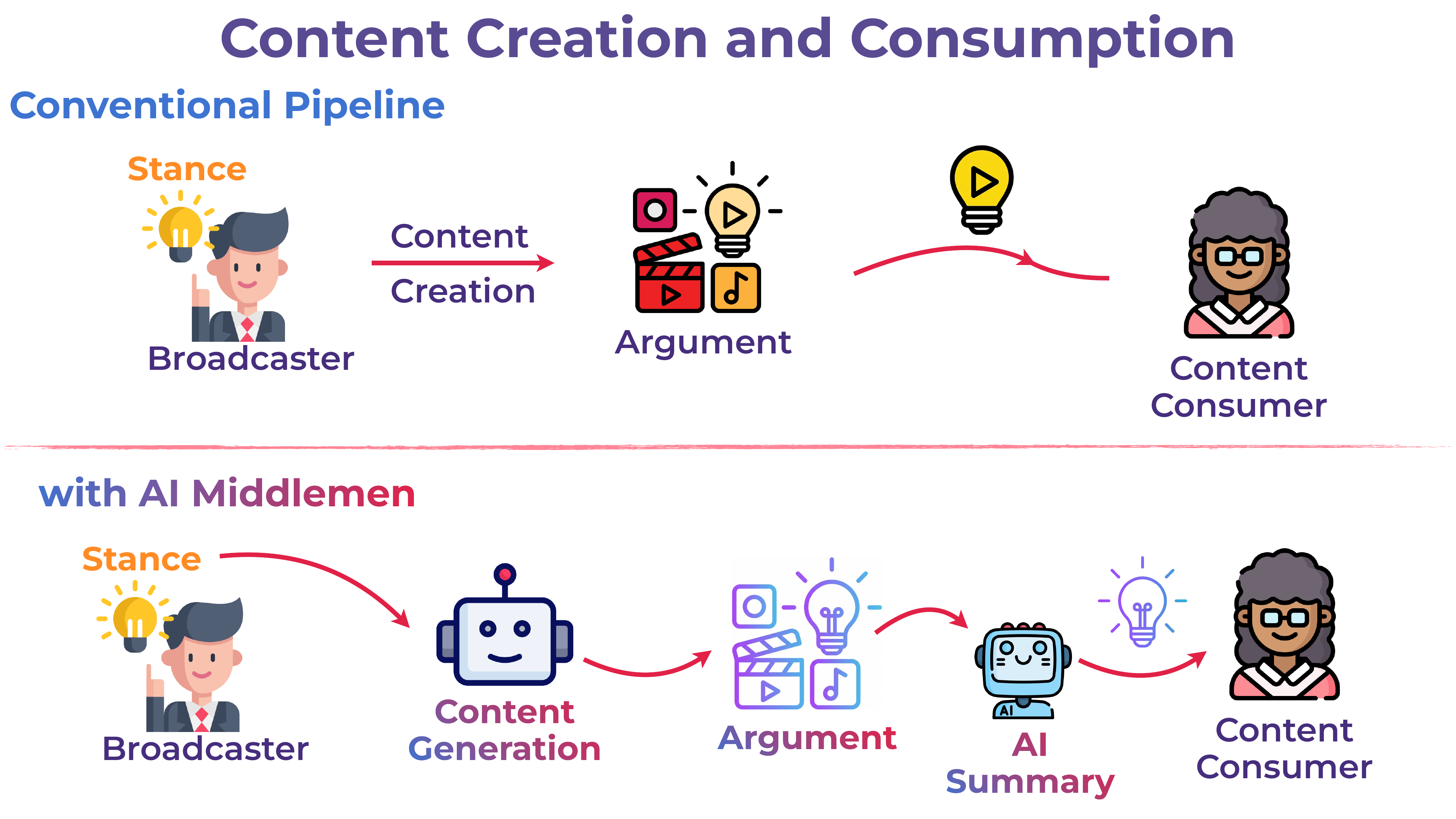}
    \caption{
        \textbf{AI-mediated communication model.}
        A sender generates an argument with LLM assistance from an initial stance; a receiver extracts the stance from the argument with LLM assistance.
    }
    \label{fig:model}
\end{figure}

\fi

To illustrate, Fig.~\ref{fig:teaser} presents the empirical transition matrices (averaged across 100 repetitions) from two experiments with OpenAI's \texttt{GPT-5.4} (complete results with standard errors in Fig.~\ref{fig:teaser-se}). For a fixed proposition, the $(i,j)$ entry of the matrix represents the probability of ending in the $j$th choice, after our proposed generation-extraction process, given the $i$th choice as the initial stance. We compare the empirical matrices of proposition ``We should increase fuel tax'' (Fig.~\ref{fig:teaser}A) and ``The use of AI should be abandoned'' (Fig.~\ref{fig:teaser}B) against the identity matrix (Fig.~\ref{fig:teaser}C), which represents the ideal of perfect preservation of each stance through generation and extraction.

We observe three major drift patterns. First, \textit{polarization} occurs when moderate inputs (``Agree''/``Disagree'') intensify to extreme positions (``Agree strongly''/``Disagree strongly'') (Fig.~\ref{fig:teaser}A). Second, \textit{deviation from neutrality} manifests as systematic shifts from neutral stances toward opinionated positions (Fig.~\ref{fig:teaser}B). Third, \textit{flipping} involves complete reversals, where positive inputs (``Agree''/``Agree strongly'') transform into negative outputs (``Disagree''/``Disagree strongly'') (Fig.~\ref{fig:teaser}B), or vice versa. Two further patterns, \textit{moderation} and \textit{moderation to neutral}, are minor (Section~\ref{sec:drift patterns}). Some propositions exhibit a single dominant pattern (Fig.~\ref{fig:teaser}A) while others display several patterns operating simultaneously across different initial stances (Fig.~\ref{fig:teaser}B).

To move beyond topic-specific behavior and establish a general measure of \textit{stance drift}, we scale our framework across diverse propositions using a debate dataset~\citep{slonim2021autonomous}. This expansion enables us to define the stance preservation rate (SPR)—a quantitative metric that reveals pervasive \textit{stance drift} across both topics and models. We further test several mitigation strategies, including in-context learning and reflection, and find that they offer limited improvement in stance preservation; raising the reasoning effort of \texttt{GPT-5.4} helps most, yet about a quarter of the transition mass still drifts.

Our framework, through proposition-conditioned transition matrices and a scalar measure of stance preservation, provides a systematic and repeatable way to assess risks in AI-mediated communication and offers a principled metric for future mitigation efforts.

\section{Materials and methods}\label{sec:methods}

\subsection{Input data preparation}\label{sec:dataset}

We take the 114 debate topics of the debate speeches dataset~\citep{slonim2021autonomous} as propositions and do not use the speeches themselves.
The topics are clearly stated but open to competing lines of reasoning, so they support plausible arguments on both sides and test whether a model preserves an intended stance rather than restates a factual claim.

\subsection{Experimental framework: generation and extraction}\label{sec:framework}

We model AI-mediated human-to-human communication as a two-step generation-extraction process (Fig.~\ref{fig:framework}). In each trial, a first LLM (Agent 1) is given a proposition along with a specified initial stance and is prompted to generate an argument. (Here, a \textit{proposition} is a declarative statement, a \textit{stance} is the attitude toward that proposition, chosen from a five-point Likert scale: Agree strongly, Agree, Neutral, Disagree, Disagree strongly as in social science surveys~\citep{johns2010likert, croasmun2011using}, and an \textit{argument} is a short passage that expresses the reasoning behind that stance.) For example, if the proposition is about AI abandonment (``The use of AI should be abandoned.''), an initial stance of ``Disagree strongly'' would lead Agent 1 to produce an argument arguing strongly against abandoning AI. A second call to the same model (Agent 2), with no access to the first call's context, then extracts the stance from Agent 1's argument by reading both the argument and the proposition, outputting one of the five stance options (see Figs.~\ref{fig:example1}--\ref{fig:example2} for individual examples).

In both the generation and extraction processes, options are presented in the prompt as multiple-choice questions, following prior work~\citep{ren2023self, durmus2023towards, hong2025hypothesis}. For LLMs that provide token-level log probabilities via API, extraction uncertainty is represented directly by the model's output probability distribution in a single trial. To account for generation-stage randomness, we repeat trials for each stance of each proposition and average the resulting extraction distributions. Finally, the transition probabilities will be estimated by the average (across trials) probabilities assigned to each of the five choices by the LLM when API-available, or the observed frequencies otherwise.
Table~\ref{tab:reproducibility} in the appendix lists, for every run, the model identifier, the run dates, and which of the two estimates was used.

\ifpagecut\else
% fig_framework.tex -- the two-step generation and extraction framework figure (main text of the preprint; appendix of the ICLR build, TASK 4.2).
\begin{figure}[t]
    \centering
    \includegraphics[width=0.8\textwidth]{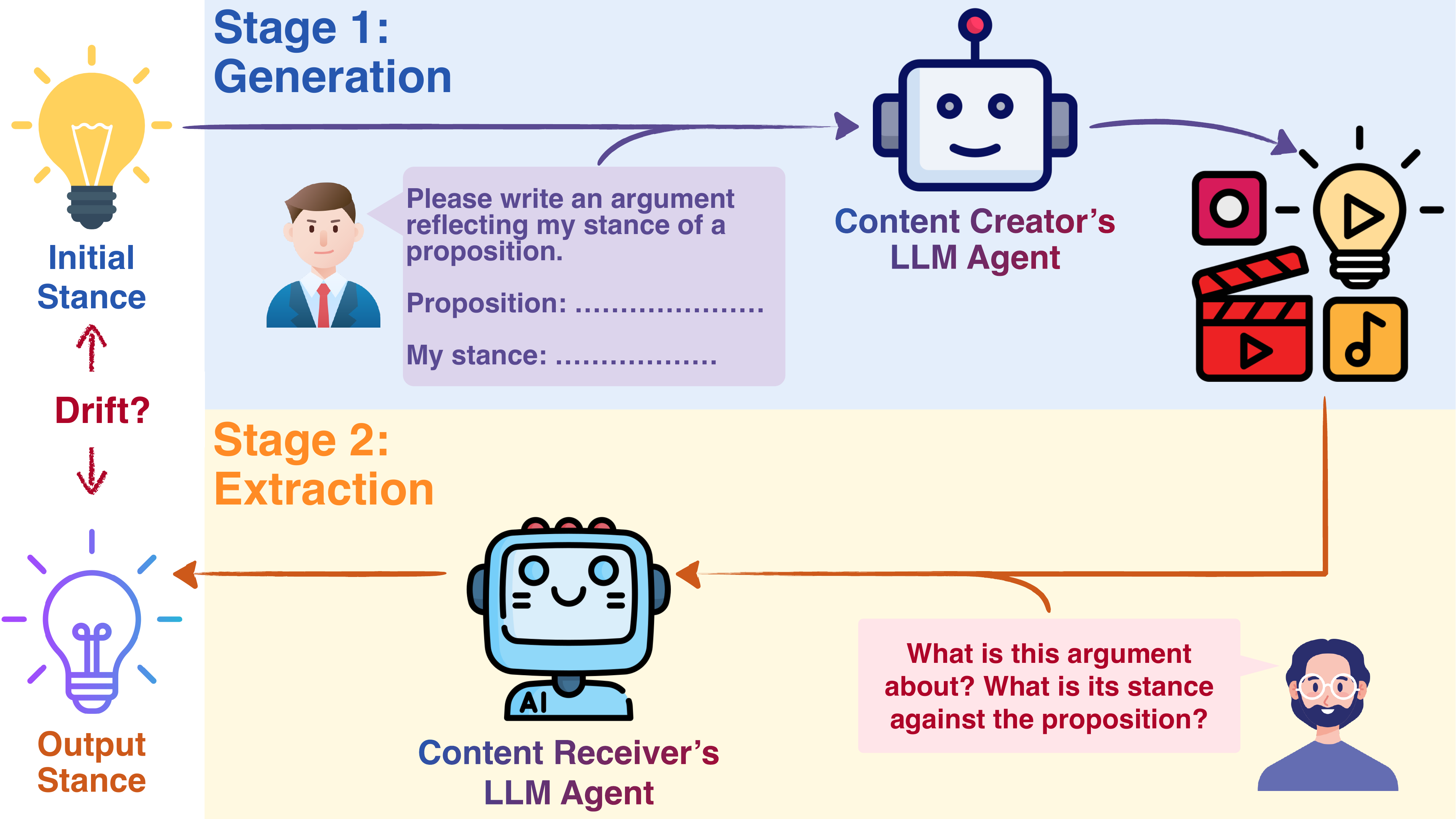}
    \caption{
        \textbf{The two-step generation and extraction framework.}
        An initial stance on a proposition is expanded into an argument, and the stance is extracted back from the argument.
    }
    \label{fig:framework}
\end{figure}

\fi

\subsection{Transition matrix and stance preservation rate (SPR) metric}\label{sec:metric}

The transition matrix resulting from this process would be a $5\times5$ matrix. Without further specification, we overload the notation and use the term ``the $i$th choice'' to refer to the corresponding stance, where $i$ ranges from 0 to 4, representing the five choices: Agree strongly, Agree, Neutral, Disagree, and Disagree strongly. The $(i,j)$ entry of the transition matrix represents the probability of returning to the $j$th choice given the $i$th choice as the initial choice. 

Ideally, the transition matrix resulting from this process would be the identity matrix, indicating complete preservation of the original stance throughout the process. Let $T$ denote the transition matrix, with entries $T_{ij}$, and $I$ the identity matrix. Since each row of $T$ sums to one, the entrywise $L_1$ distance between $T$ and $I$ equals $\sum_{i,j}|T_{ij}-I_{ij}| = 2\sum_{i=0}^{4}(1-T_{ii})$, which depends only on the diagonal elements of $T$. We therefore examine the diagonal elements, which represent the probabilities of returning to the original stance after the generation-extraction.

We conduct our experiments on 112 of the 114 topics from the debate speeches dataset, reserving the remaining two topics as examples for in-context learning, as discussed in Section~\ref{sec:mitigation}. Each topic undergoes 100 repeated trials, yielding empirical $5\times5$ transition matrices whose diagonal elements $(i,i)$ quantify stance preservation: the probability that the extraction stage recovers the same stance supplied to the generation stage. We define this diagonal probability as the ``success rate'' and estimate it by averaging across trials. All raw data and estimated transition matrices are available in our code repository.

To quantify stance preservation across the diverse topics in the debate speeches dataset, we define the stance preservation rate (SPR) as the average success rate across all topics and initial stances—that is, the fraction of messages whose original stance survives the generation-extraction process (unless otherwise specified, SPR assumes a uniform prior over initial stances; alternative priors can be incorporated by weighted averaging over diagonal transition probabilities). This metric provides a single, interpretable measure of how faithfully models maintain communicative intent, and thus offers a clear benchmark for comparing model reliability in AI-mediated communication and assessing mitigation strategies.

\ifpagecut\else
% spr_extensions.tex -- the SPR-extensions paragraph of the methods (main text of the preprint; appendix A of the ICLR build, TASK 4.2).
SPR is the complement of the expected 0--1 loss under a uniform prior over initial stances. Any other prior, or any other loss over stance pairs such as one weighted by distance on the Likert scale, is a weighted sum of the entries of the same transition matrix and requires additional assumptions on the weights. These extensions share the same transition-matrix foundation; the strict SPR used throughout this paper represents a general baseline with no additional assumptions.

\fi

\subsection{Stance drift patterns}\label{sec:drift patterns}

Off-diagonal entries of the transition matrix record \textit{stance drift}.
We partition them into five patterns defined over the index pairs $(i,j)$ with $i\neq j$.
\textit{Polarization} is the pair set $\{(1,0),(3,4)\}$: a moderate stance is extracted as the extreme stance on the same side.
\textit{Moderation} is the pair set $\{(0,1),(4,3)\}$: an extreme stance is extracted as the moderate stance on the same side.
\textit{Moderation to neutral} is the pair set $\{(i,2): i\neq 2\}$: a non-neutral stance is extracted as Neutral.
\textit{Flipping} is the pair set $\{(i,j): i\in\{0,1\},\, j\in\{3,4\}\}\cup\{(i,j): i\in\{3,4\},\, j\in\{0,1\}\}$: a stance is extracted as a stance on the opposite side.
\textit{Deviation from neutrality} is the pair set $\{(2,j): j\neq 2\}$: Neutral is extracted as a non-neutral stance.
The five sets are disjoint and together cover every off-diagonal pair, so each off-diagonal entry belongs to exactly one pattern.
The mass of a pattern for initial stance $i$ is the sum of the transition probabilities $(i,j)$ over the pairs of that pattern with that $i$.
The five pattern masses for initial stance $i$ sum to one minus the success rate of $i$.
As with SPR, the overall mass of a pattern is the average of its per-stance masses over the five initial stances.
We call one minus SPR the stance drift rate (SDR).
The five overall pattern masses are the components of the SDR and sum to it.

\subsection{Mitigation attempts}\label{sec:mitigation}

Beyond measuring \textit{stance drift}, we ask whether straightforward interventions can restore fidelity in AI-mediated communication. Throughout, the default configuration denotes a run with the default prompts (Appendix~\ref{sec:prompts}). We tested four targeted strategies with \texttt{GPT-4o-mini} (full details in the appendix), each addressing a distinct hypothesized failure mode. (i) Multiple extraction addresses \textit{positional bias}, the documented tendency of LLMs to favor options at specific positions~\citep{pezeshkpour2024order,zheng2024mcq}: by running extraction five times with randomly shuffled option orderings and averaging the resulting probability distributions, we sought to reduce systematic position effects. (ii) In-context learning supplies three high-quality examples during generation and extraction, aiming to anchor model behavior through demonstration. (iii) Assertion passes the literal stance options to the extraction prompt and asks the model to choose one of them first and then map it to a letter, with the aim of directing the model's attention to the stance labels. (iv) Reflection adds one instruction to each of the two prompts: Agent 1 is asked to compare its drafted argument against the specified stance and to revise it if the two do not align, and Agent 2 is asked to compare its selected option against the argument and to revise the choice likewise, with the aim of catching drift before the output is returned.

\section{Results}\label{sec:results}
\subsection{Fidelity gap across models}
\begin{figure}[ht]
    \centering
    \includegraphics[width=.9\textwidth]{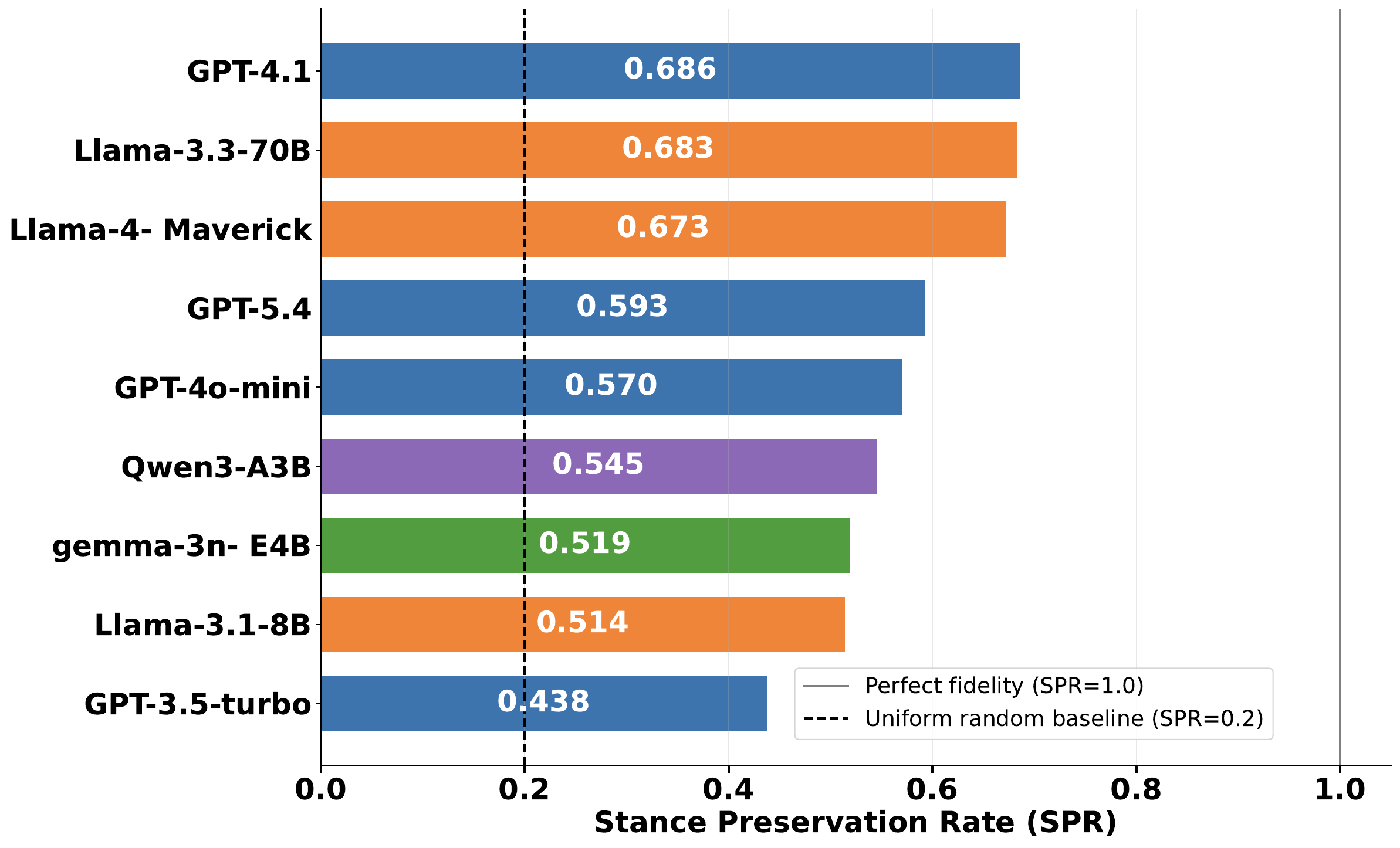} 
    \caption{
        \textbf{SPR across models under the default configuration.}
		The solid rule marks perfect fidelity (SPR $=1$) and the dashed rule the uniform random baseline (SPR $=0.2$). % lint:ok
    }
    \label{fig:metric}
\end{figure}

We evaluated nine LLMs on the debate-speeches dataset (Fig.~\ref{fig:metric}). None of the models exceeded an SPR of $0.7$ under the default configuration (Fig.~\ref{fig:metric}). In this model set, higher SPR tended to appear in larger models (e.g., \texttt{GPT-4.1}, \texttt{Llama-3.3-70B}, \texttt{Llama-4-Maverick}) (Fig.~\ref{fig:metric}), although these comparisons are confounded by architecture and training differences.  % lint:ok
For example, \texttt{GPT-5.4} under the default configuration has a lower SPR ($\SPRGPTfivefour$) than \texttt{GPT-4.1} ($\SPRGPTfourone$), whereas with third-person reflection and medium reasoning effort it reaches $\SPRGPTfivefourThirdPersonReflectionMedium$ (Section~\ref{sec:robustness}).

\subsection{Stance-level heterogeneity in preservation failure}
A stance-level analysis further reveals substantial heterogeneity in preservation performance (Table~\ref{tab:rates of preservation and bonferroni corrected CI inclusion}A).
Although most models achieve high success rates for specific initial stances (e.g., ``Agree strongly'' for \texttt{GPT-4o-mini} and ``Neutral'' for \texttt{gemma-3n-E4B}), every model except \texttt{Llama-4-Maverick} (lowest stance: $\SuccessRateDisagreeStronglyLlamaFourMaverick$) falls below $0.3$ for at least one stance (e.g., ``Disagree'' for \texttt{GPT-4.1} and \texttt{Llama-3.3-70B}). % lint:ok
A consistent cross-model pattern emerges: with the exception of \texttt{Llama-3.1-8B} and \texttt{gemma-3n-E4B}, the highest success rates are obtained for the initial stance ``Agree strongly'', exceeding $0.85$ for all but \texttt{GPT-3.5-turbo}. % lint:ok
In contrast, neutrality is less reliably preserved; only \texttt{gemma-3n-E4B}, \texttt{Llama-3.3-70B}, and \texttt{GPT-4.1} reach $0.8$ for the initial stance ``Neutral''. % lint:ok
For all models but \texttt{Llama-4-Maverick}, the lowest success rate falls on one of the moderate stances ``Agree'' and ``Disagree'' (e.g., ``Agree'': $\SuccessRateAgreeGPTfivefour$ and ``Disagree'': $\SuccessRateDisagreeGPTfivefour$ for \texttt{GPT-5.4}).

\ifpagecut\else
% tab_success_ci.tex -- Table 1 of the preprint: success rates and Bonferroni CI inclusion for the nine models (appendix of the ICLR build, TASK 4.2).
\begin{table}[h!]
  \centering
  \caption{
        \textbf{Stance preservation by initial stance for the nine models under the default configuration.}
        (\textbf{A}) Success rates.
        (\textbf{B}) Proportion of propositions whose one-sided 99\% CI of the success rate, with Bonferroni correction, includes 1.} % lint:ok
  \begin{minipage}[b]{\textwidth}
    \scriptsize
    \centering
    \vspace{1em}
    (\textbf{A})\\
    % Success rate (mean diagonal probability) by initial stance
\begin{tabular}{lccccc}
		\\
		\hline
		Model & Agree strongly & Agree & Neutral & Disagree & Disagree strongly \\
		\hline
		\texttt{GPT-4o-mini} & .933 (.021) & .108 (.016) & .717 (.026) & .310 (.035) & .784 (.032) \\
		\texttt{GPT-4.1} & .962 (.014) & .599 (.031) & .800 (.026) & .168 (.027) & .903 (.016) \\
		\texttt{GPT-5.4} & .996 (.002) & .147 (.022) & .703 (.032) & .136 (.023) & .982 (.010) \\
		\texttt{GPT-3.5-turbo} & .621 (.031) & .484 (.031) & .313 (.023) & .244 (.027) & .527 (.036) \\
		\texttt{gemma-3n-E4B} & .595 (.040) & .348 (.034) & .920 (.019) & .114 (.022) & .617 (.041) \\
		\texttt{Llama-3.3-70B} & .954 (.015) & .505 (.036) & .851 (.023) & .161 (.026) & .942 (.017) \\
		\texttt{Llama-3.1-8B} & .368 (.019) & .768 (.013) & .442 (.014) & .104 (.007) & .887 (.011) \\
		\texttt{Llama-4-Maverick} & .857 (.024) & .674 (.031) & .675 (.032) & .642 (.037) & .515 (.042) \\
		\texttt{Qwen3-A3B} & .939 (.020) & .268 (.030) & .402 (.034) & .441 (.041) & .677 (.039) \\
		\hline
\end{tabular}

        \vspace{1em} % vertical space between subfigures
    \end{minipage}
  % Second subfigure (bottom)
\begin{minipage}[b]{\textwidth}
    \scriptsize
    \centering
    (\textbf{B})\\
    % Bonferroni-Corrected CI Inclusion Rate
\begin{tabular}{lccccc}
		\\
		\hline
		Model & Agree strongly & Agree & Neutral & Disagree & Disagree strongly \\
		\hline
		\texttt{GPT-4o-mini} & .223 (.040) & .000 (.000) & .402 (.047) & .116 (.030) & .536 (.047) \\
		\texttt{GPT-4.1} & .920 (.026) & .286 (.043) & .616 (.046) & .054 (.021) & .732 (.042) \\
		\texttt{GPT-5.4} & .991 (.009) & .027 (.015) & .527 (.047) & .018 (.013) & .973 (.015) \\
		\texttt{GPT-3.5-turbo} & .000 (.000) & .045 (.020) & .018 (.013) & .000 (.000) & .080 (.026) \\
		\texttt{gemma-3n-E4B} & .348 (.045) & .098 (.028) & .500 (.047) & .009 (.009) & .464 (.047) \\
		\texttt{Llama-3.3-70B} & .920 (.026) & .295 (.043) & .714 (.043) & .054 (.021) & .902 (.028) \\
		\texttt{Llama-3.1-8B} & .000 (.000) & .241 (.041) & .000 (.000) & .000 (.000) & .643 (.045) \\
		\texttt{Llama-4-Maverick} & .705 (.043) & .357 (.045) & .482 (.047) & .455 (.047) & .393 (.046) \\
		\texttt{Qwen3-A3B} & .607 (.046) & .071 (.024) & .107 (.029) & .321 (.044) & .420 (.047) \\
		\hline
\end{tabular}

  \end{minipage}
  \label{tab:rates of preservation and bonferroni corrected CI inclusion}
\end{table}

\fi

To further assess whether any observed preservation could be consistent with perfect fidelity, we conduct a confidence interval (CI) analysis for the success rate (Table~\ref{tab:rates of preservation and bonferroni corrected CI inclusion}B). For each proposition and initial stance, we estimate the success rates over 100 trials and construct a one-sided 99\% CI using the central limit theorem. We consider stance preservation statistically indistinguishable from perfect fidelity if the CI includes $1$. Because we perform this test across 112 propositions for each initial stance, we apply Bonferroni correction to widen each CI to account for multiple comparisons and report the proportion of propositions whose CIs include $1$. This proportion is below $0.65$ in $\BonferroniBelowCells$ of the 45 cells (Table~\ref{tab:rates of preservation and bonferroni corrected CI inclusion}B). Even under this conservative Bonferroni correction—which tends to enlarge intervals and increase inclusion rates—the low proportions confirm that LLMs consistently fail to reliably recover initial stances.  % lint:ok

\subsection{Distribution of stance drift across patterns}

Table~\ref{tab:drift patterns} partitions the SDR of each model into the five drift patterns defined in Section~\ref{sec:drift patterns}.
Polarization is the largest pattern for seven of the nine models; for \texttt{GPT-3.5-turbo} and \texttt{gemma-3n-E4B}, flipping is the largest.
Polarization, deviation from neutrality, and flipping together carry most of the drift of every model, and we refer to them as the three major patterns; moderation and moderation to neutral are minor.
Polarization exceeds moderation for all nine models, which a symmetric difficulty in reading intensity would not produce.

\begin{table}[h!]
  \centering
  \caption{
        \textbf{SPR and the components of the SDR by drift pattern for the nine models under the default configuration.}
        Mean (SE) over the 112 propositions of the per-proposition values averaged over the five initial stances; each row sums to 1 up to rounding.
        Per-stance values and all tested models: Tables~\ref{tab:sdr summary}--\ref{tab:sdr deviation} in the appendix.}
  \scriptsize
  \resizebox{\textwidth}{!}{% SDR Summary: rows=models, columns=[SPR, 5 drift patterns = components of SDR (1 - SPR)]; fractions of the transition mass, each row sums to 1. Stage-1: mean over 5 initial stances per (topic, perm). Stage-2: mean (SE) across (topic, perm) samples.
\begin{tabular}{lcccccc}
		\\
		\hline
		Model & SPR & Polarization & Moderation & Moderation to neutral & Flipping & Deviation from neutrality \\
		\hline
		\texttt{GPT-4o-mini} & .570 (.009) & .246 (.011) & .022 (.004) & .009 (.002) & .097 (.012) & .057 (.005) \\
		\texttt{GPT-4.1} & .686 (.007) & .244 (.005) & .024 (.004) & .001 (.000) & .005 (.001) & .040 (.005) \\
		\texttt{GPT-5.4} & .593 (.008) & .328 (.006) & .002 (.001) & .000 (.000) & .018 (.004) & .059 (.006) \\
		\texttt{GPT-3.5-turbo} & .438 (.008) & .158 (.009) & .087 (.006) & .000 (.000) & .180 (.014) & .137 (.005) \\
		\texttt{gemma-3n-E4B} & .519 (.013) & .165 (.010) & .041 (.006) & .082 (.010) & .178 (.015) & .016 (.004) \\
		\texttt{Llama-3.3-70B} & .683 (.011) & .191 (.007) & .014 (.003) & .015 (.003) & .069 (.009) & .030 (.005) \\
		\texttt{Llama-3.1-8B} & .514 (.006) & .174 (.004) & .085 (.004) & .012 (.001) & .103 (.010) & .112 (.003) \\
		\texttt{Llama-4-Maverick} & .673 (.012) & .122 (.007) & .057 (.006) & .005 (.001) & .078 (.009) & .065 (.006) \\
		\texttt{Qwen3-A3B} & .545 (.010) & .216 (.011) & .065 (.008) & .001 (.001) & .052 (.009) & .120 (.007) \\
		\hline
\end{tabular}
}
  \label{tab:drift patterns}
\end{table}

\subsection{Stance drift survives robustness checks and is only partly alleviated by mitigation}\label{sec:robustness}

Additional robustness checks that relax the perfect-preservation criterion by introducing thresholds strictly below $1$ consistently confirm the presence of \textit{stance drift} (Table~\ref{tab:pass rate} and Appendix~\ref{sec:pass rate}). To avoid fixing a single threshold, we compute the area under the curve (AUC) of the pass rate, the proportion of proposition--initial stance combinations whose success rate exceeds the threshold, as thresholds vary from $0.5$ to $1$, yielding a single summary statistic which provides similar results as SPR regarding model comparison. Results show that models consistently underperform across this continuum, reinforcing that \textit{stance drift} is not an artifact of requiring a success rate of exactly $1$ (Fig.~\ref{fig:AUC}). % lint:ok

We conducted ablation analyses using \texttt{GPT-4o-mini} (default configuration, SPR $=\SPRGPTfouromini$) to assess potential confounds in the generation-extraction pipeline (Table~\ref{tab:success rates}, Fig.~\ref{fig:SPR_all_models}). First, to test whether the temperature (the parameter that controls randomness when inferring the next token) of the extraction stage drives the low SPR, we re-analyzed the raw results with each trial's extraction vector replaced by a point mass on its most probable option, which is equivalent to setting the temperature to $0$ during extraction. This re-analysis yielded an SPR of $\SPRGPTfourominiTempZero$, indicating no meaningful change. Second, to examine positional and formatting biases, we reversed the choice order (negative-to-positive), obtaining a comparable SPR of $\SPRGPTfourominiReversed$. We further implemented the multiple extraction protocol (Section~\ref{sec:mitigation}) to mitigate potential positional bias; this procedure likewise produced no substantive improvement (SPR $=\SPRGPTfourominiMultipleExtraction$). Third, we evaluated prompt-level mitigation strategies. In-context learning resulted in an SPR of $\SPRGPTfourominiInContext$, assertion yielded $\SPRGPTfourominiAssert$, and reflection yielded $\SPRGPTfourominiReflection$, none differing meaningfully from the default configuration.

Reflection is the one strategy we also tested on \texttt{GPT-5.4}, in three configurations that each add one change to the previous one: reflection; third-person reflection, which phrases the instruction from the perspective of an impartial third-party reviewer; and third-person reflection with medium reasoning effort (full details in the appendix).
The SPR rises from $\SPRGPTfivefour$ to $\SPRGPTfivefourReflection$, $\SPRGPTfivefourThirdPersonReflection$ and $\SPRGPTfivefourThirdPersonReflectionMedium$, and polarization falls at every step, from $\PolMassGPTfivefour$ to $\PolMassGPTfivefourThirdPersonReflectionMedium$ (Fig.~\ref{fig:gpt54 configurations} and Table~\ref{tab:sdr summary} in the appendix).
The two reflection steps move part of the drift of the initial stance ``Agree'' from polarization into flipping (from $\FlipMassAgreeGPTfivefour$ to $\FlipMassAgreeGPTfivefourThirdPersonReflection$).
The reasoning step removes most of that flipping ($\FlipMassAgreeGPTfivefourThirdPersonReflectionMedium$) and gives the largest gain, but lowers the success rate for ``Neutral'' from $\SuccessRateNeutralGPTfivefourThirdPersonReflection$ to $\SuccessRateNeutralGPTfivefourThirdPersonReflectionMedium$.
Even in this last configuration, about a quarter of the transition mass still drifts and polarization remains the largest pattern.

\textit{Stance drift} is therefore not fully explained by the prompting and decoding factors tested here.

\subsection{An exploratory human-extraction study locates drift at both stages}\label{sec:human study}

The ablation analyses above treat the generation-extraction pipeline as a whole. To probe whether drift originates in generation, extraction, or both, we conducted a targeted human-extraction study (Appendix~\ref{sec:human study details}) on the proposition ``The use of AI should be abandoned'' using 100 arguments generated by \texttt{GPT-5.4} (20 per initial stance). Nine human annotators, working without AI assistance, each labeled all 100 arguments via a structured survey; their majority vote served as the reference label throughout.

Comparing the human-majority against the known initial stance reveals generation-stage drift; comparing the AI-majority against the human-majority reveals extraction-stage drift. The results reveal that both stages contribute to \textit{stance drift} yet show a systematic pattern by initial stance (Table~\ref{tab:human extraction}; one example argument per drift pattern in Table~\ref{tab:drift examples}). For ``Agree'', flipping to ``Disagree'' was already present across all 20 generated arguments, implicating generation-stage drift. For ``Disagree'', polarization toward ``Disagree strongly'' was absent from human-majority labels but appeared consistently in AI extraction, implicating the extraction stage. Deviation from neutrality showed a mixed pattern: 5 of 20 ``Neutral''-initiated arguments were already non-neutral (generation-stage drift), and 13 deviated at AI extraction. 

\ifpagecut\else
% human_ai_gap.tex -- the human-AI agreement sentences of the human-study paragraph (main text of the preprint; appendix A of the ICLR build, TASK 4.2).
To quantify the human–AI performance gap at the extraction stage, we computed, for each individual trial (each single human annotation and each single AI extraction vote), the agreement with the human-majority label. This comparison serves to validate both the task design and the feasibility of high-fidelity extraction. Note that this validation is non-trivial: even under the majority-vote reference, the minimum expected per-annotator agreement is bounded below by $2/9 \approx \HumanExtractionChanceFloor$. Human annotators achieved $\HumanExtractionAgreement$ (SE $\HumanExtractionAgreementSE$), well above this floor, confirming that the arguments carry recoverable stance signals and that the task admits substantially better-than-chance extraction. The AI shortfall, \texttt{GPT-5.4} at $\AIExtractionAgreement$ (SE $\AIExtractionAgreementSE$), is therefore consistent with an extraction deficit and not accounted for by ambiguity in the arguments.
\fi
Given the intensive demands of unaided human labeling, this study is limited to a single proposition and a small annotator pool; online deployment on platforms such as Amazon Mechanical Turk would risk contamination from AI-assisted responses, undermining the human-only baseline. It is intended as a proof-of-concept decomposition and human benchmark rather than a comprehensive measurement.

\section{Implications and open questions}\label{sec:discussion}
\subsection{Sources of stance drift}
\textit{Stance drift}, the failure of LLMs to preserve original positions, emerged consistently across the models and propositions examined here and was only partly alleviated by the decoding, format and prompt changes we tested (Section~\ref{sec:robustness}).
These results motivate treating stance fidelity as a distinct evaluation target in AI-mediated communication.

Our conclusion is descriptive: it characterizes \textit{stance drift} on the tested dataset and does not isolate a causal source.

As an initial step toward understanding potential sources of drift, inspection of intermediate generated arguments suggests several recurring features. First, models sometimes mishandle negation-heavy reasoning (e.g., cases in which a proposition containing negated statement is discussed in a negative stance), leading to flipping during either generation or extraction. Second, models may conflate a cautiously phrased supporting argument—written to acknowledge counterarguments or avoid overstatement—with a genuinely opposing stance. We also observed that propositions containing terms such as ``ban'' frequently coincided with flipping or misclassification.
The structure of the drift patterns (Table~\ref{tab:drift patterns}) suggests that future work should examine whether different patterns are associated with distinct sources, such as training data biases and architectural limitations in maintaining semantic invariants through paraphrase.
These observations are qualitative and specific to the policy-oriented, sometimes negation-heavy propositions of the debate speeches dataset.

\subsection{Stance preservation as a benchmark for AI-mediated communication}
As for potential future work, although the mitigation strategies tested here yielded limited gains, the framework points to a practical optimization target: stance preservation itself. In particular, SPR measured on datasets such as the debate speeches dataset, could be used as a reward in reinforcement learning to encourage LLMs to maintain stance across generations. More broadly, the generation-extraction framework can be extended beyond stance to subtler attributes such as tone, framing, and contextual emphasis. A particularly promising direction is multi-hop communication chains, where messages pass through multiple LLMs or human--AI hybrids: modelled as Markov chains, they would show whether drift accumulates over successive transmissions and whether stationary distributions mark stable distortion endpoints.

For LLM development, SPR provides a communication-fidelity benchmark that complements traditional accuracy-oriented metrics by testing whether models preserve intended positions, not only whether they produce plausible text. For social-science and opinion-sensitive research settings, our results raise caution about using current AI agents as substitutes~\citep{qin2024aiturk} for human subjects when faithful preservation of stated positions is essential. In domains such as journalism, public policy, scientific research, education, and everyday communication, reliance on LLMs for drafting, summarizing, or mediating messages may inadvertently amplify, shift, or distort stances, affecting decision-making, fairness, and public perception~\citep{acerbi2023large, liu2024p3sum, wang2025echoes, kelly2025understanding}.

% statements.tex -- the ethics and reproducibility statements of the ICLR 2027 template (TASK 4.5,
% 4.6), input by both drivers after the discussion. The ethics statement follows the seed in
% PNAS_main/coverletter.tex (DECISIONS.md 2026-09-09); every other fact is taken from the
% appendix. \CodeURL is defined by the driver (the anonymous mirror in ICLR_main/iclr.tex, the
% GitHub repository in ICLR_main/arxiv.tex).
\subsection*{Ethics statement}
The human-extraction study (Section~\ref{sec:human study} and Appendix~\ref{sec:human study details}) collected stance labels from nine annotators.
The annotators were informed of the nature and purpose of the task prior to participation, and their involvement was voluntary.
All other data in this work are the public debate propositions of \citet{slonim2021autonomous} and text generated from them by the models under study.
During the course of this work, one author became a visiting researcher at the organization that develops one of the evaluated model families.
That organization had no role in the design, execution or reporting of the study.

\subsection*{Reproducibility statement}
Table~\ref{tab:reproducibility} in Appendix~\ref{sec:reproducibility details} lists, for every run, the model identifier requested through the API, the API provider, the prompt, the run dates and whether the transition probabilities were estimated from token-level log probabilities or from observed frequencies.
The prompts of every configuration are given verbatim in Appendix~\ref{sec:prompts} and in the appendix subsections on the mitigation strategies.
The code, the raw per-trial extraction distributions and the estimated transition matrices of every run, the human-study labels, and the scripts that produce every table and figure of this paper are available at \CodeURL.

% ai_use.tex -- the AI use statement (TASK 4.8, operator's text of 2026-09-14; the task phrases
% follow the ICLR 2027 AI Policy for Authors). Input by both drivers: iclr.tex and, since the
% review round of 2026-09-14 (C17), arxiv.tex.
\subsection*{AI use statement}
In this work, we used generative AI tools to generate synthetic data sets, design or provide feedback on research methodology or experiments, implement methods, clean and reformat dataset, and support qualitative and thematic data analysis.
We have not used generative AI tools to help develop theoretical models or conceptual frameworks, propose or refine hypotheses, or interpret results, and assisting with translation, formulating mathematical claims, providing critical ingredients for proving mathematical claims, and assisting in the writing of proofs are not applicable to this work.
Additionally, we used generative AI tools to search the literature; to create or edit code, namely the scripts that run the experiments, aggregate the results, and draw the figures and tables from the results; to draft parts of the paper; to edit the paper to improve readability; and to format references.
We have reviewed all AI-assisted work.
The synthetic data are the arguments and stance extractions produced by the language models under study; they are the object of measurement, and the pipeline that produces them is described in Section~\ref{sec:methods} and the appendix.
AI-assisted code was reviewed and run by the authors.
AI-assisted text was reviewed and revised by the authors.
We take responsibility for the final content of this work, including text, claims or artifacts produced with the aid of generative AI.

% acknowledgments.tex -- acknowledgments, funding, competing interests and data availability,
% carried over from the PNAS version (sections/backmatter.tex and sections/frontmatter.tex).
% Input by the preprint driver (ICLR_main/arxiv.tex) only; the ICLR submission is anonymous
% (TASK 4.1b, 4.3). \CodeURL is defined by the driver.
\subsubsection*{Acknowledgments}
We thank Jie~Ren and Deqing~Fu for helpful discussions, assistance in experiments and feedback on earlier drafts of this manuscript.
We thank Xiaolong~Wang, Qiannan~Huang, Shiqi~Lei, John~Park, Yifan~Zhang and Helen Miao~Yu for participating in human extraction experiment as annotators.
X.~T. is funded by start-up fund at HKU and USC Marshall STARS.
While J.~B. did this work as part of his employment by USC, during the course of the work he became a Visiting Academic Researcher at Google, which developed the Gemma LLMs.

\subsubsection*{Data availability}
The code repository as well as the results for our experiments on stance preservation in large language models (LLMs) is available on GitHub, at \CodeURL.

\bibliographystyle{iclr2027_conference}
\bibliography{pnas-sample}

\appendix
% appendix.tex -- the supporting information of the PNAS version, ported as the appendix of the
% ICLR port (TASK 4.1). Figures and tables are numbered S1, S2, ... as in the SI so that the
% main-text references keep their form. Input after \appendix by both drivers.
\setcounter{figure}{0}\renewcommand{\thefigure}{S\arabic{figure}}
\setcounter{table}{0}\renewcommand{\thetable}{S\arabic{table}}

%%%%%%%%%%%%%%%% MATERIALS AND METHODS %%%%%%%%%%%%%%%

\section{Additional materials and methods}\label{app:methods}

\subsection{Algorithm of proposed experimental framework: generation and extraction}
\begin{enumerate}
    \item \textbf{Input:} Proposition $s$, LLM model $\mathcal{M}$, generation prompt $F_{para}$, extraction prompt $F'$, the total number of repeated trials $K$.
    \item For $k = 1$ to $K$, repeat:
    \begin{enumerate}
        \item For initial choice $X_0=i \in \{0,1,2,3,4\}$, repeat:
        \begin{enumerate}%[label=(\roman*)]
            \item Convert $X_0$ to the corresponding choice $choice(X_0) \in$ \{`A', `B', `C', `D', `E'\}.
            \item Compose the generation prompt with the proposition and the initial choice,\\$F_{para}(s,choice(X_0))$, feed to $\mathcal{M}$, obtain output argument $z_{X_0}$.
            \item Compose the extraction prompt with the proposition and the argument $F'(s, z_{X_0})$, feed to $\mathcal{M}$ without any history, obtain output softmax probability scores $P_k(X_1=j \mid X_0=i)$, where $j \in \{0,1,2,3,4\}$ is the returned choice.
        \end{enumerate}
    \end{enumerate}
    \item Compute the estimated transition matrix entry as the average over repeated trials:
        \[\hat{\mathcal{P}}(X_1=j \mid X_0=i) = \frac{1}{K} \sum_{k=1}^{K} P_k(X_1=j \mid X_0=i).\]
        A trial whose selection step returns no valid choice is excluded from the average, and $K$ counts the remaining trials for that entry.
        This occurred for \FailedEncodeCellsWord{} trials across all runs (one for \texttt{Llama-3.1-8B}, five for \texttt{Llama-4-Maverick}).
    \item \textbf{Output:} Estimated transition matrix where the $i,j$ entry is $\hat{\mathcal{P}}(X_1=j \mid X_0=i)$.
\end{enumerate}
\ifpagecut
% fig_model.tex -- the AI-mediated communication model figure (main text of the preprint; appendix of the ICLR build, TASK 4.2).
\begin{figure}[t]
    \centering
    \includegraphics[width=0.9\textwidth]{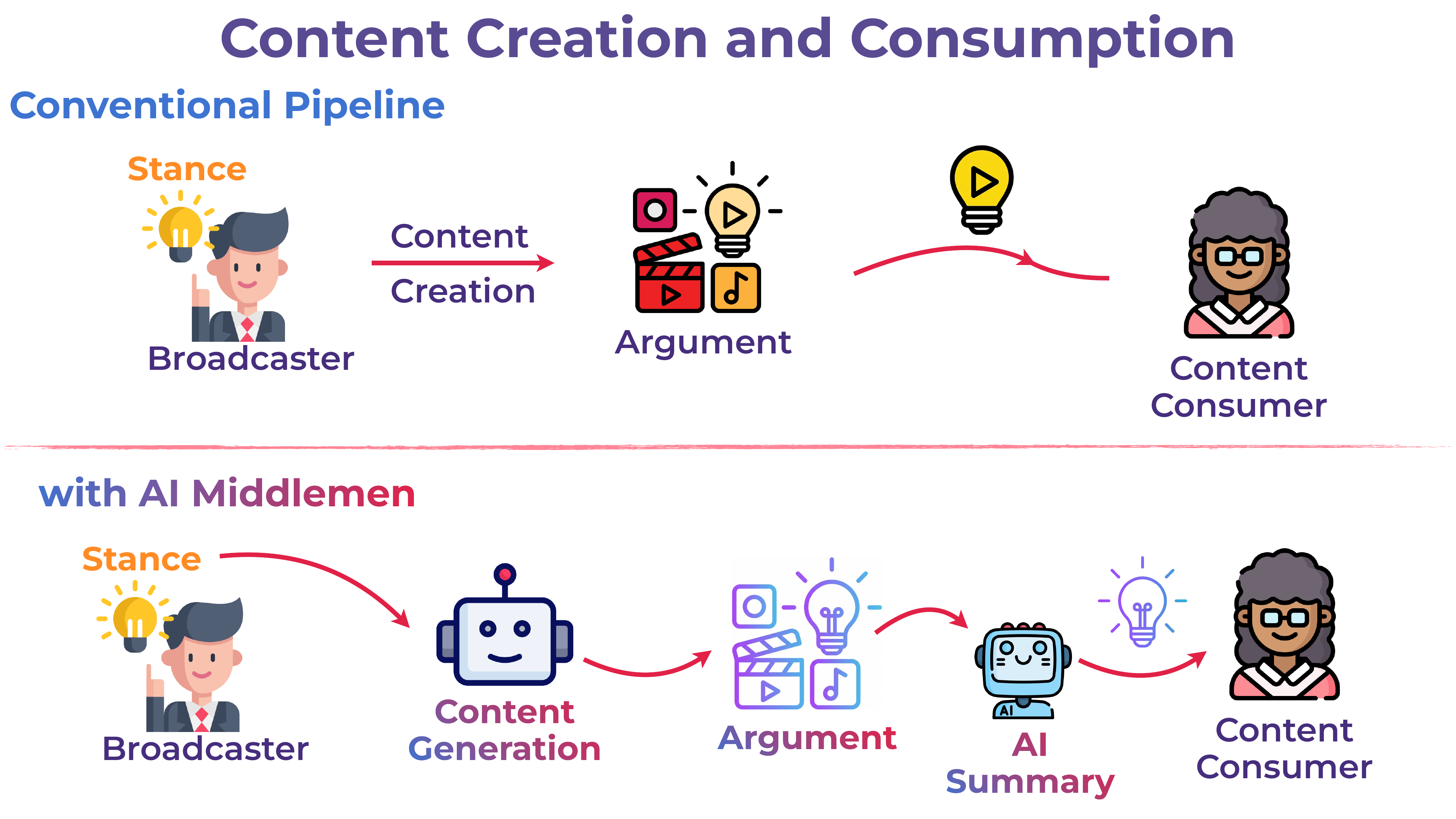}
    \caption{
        \textbf{AI-mediated communication model.}
        A sender generates an argument with LLM assistance from an initial stance; a receiver extracts the stance from the argument with LLM assistance.
    }
    \label{fig:model}
\end{figure}

% fig_framework.tex -- the two-step generation and extraction framework figure (main text of the preprint; appendix of the ICLR build, TASK 4.2).
\begin{figure}[t]
    \centering
    \includegraphics[width=0.8\textwidth]{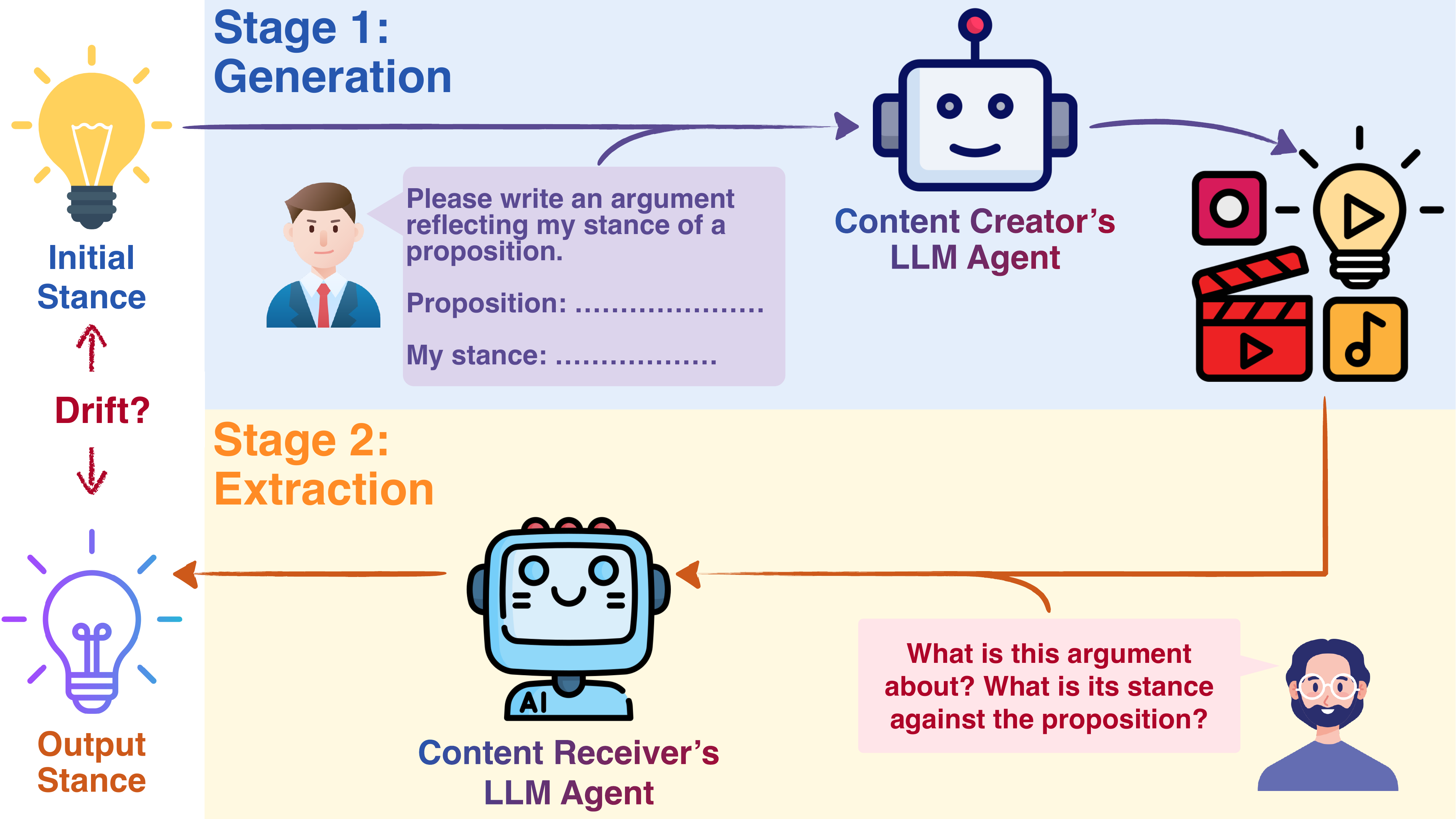}
    \caption{
        \textbf{The two-step generation and extraction framework.}
        An initial stance on a proposition is expanded into an argument, and the stance is extracted back from the argument.
    }
    \label{fig:framework}
\end{figure}

\fi
\ifpagecut
\subsection{Extensions of SPR}
% spr_extensions.tex -- the SPR-extensions paragraph of the methods (main text of the preprint; appendix A of the ICLR build, TASK 4.2).
SPR is the complement of the expected 0--1 loss under a uniform prior over initial stances. Any other prior, or any other loss over stance pairs such as one weighted by distance on the Likert scale, is a weighted sum of the entries of the same transition matrix and requires additional assumptions on the weights. These extensions share the same transition-matrix foundation; the strict SPR used throughout this paper represents a general baseline with no additional assumptions.

\fi

\subsection{Adapted debate speeches dataset} 
Notably, each debate topic in the dataset appears multiple times, corresponding to speeches generated from different sources (e.g., human expert, large language model, IBM's automated Project Debater system). For example, the topic ``We should subsidize vocational education'' appears in seven separate rows, each associated with a distinct source. In our experiments, we disregard source-specific duplicates and retain only the unique topics, which we treat as propositions. An illustrative example is shown below:
\begin{verbatim}
    {
        "topic_id":"3716",
        "topic":"We should subsidize vocational education"
    }
\end{verbatim}

\subsection{Prompts used in the experiment}\label{sec:prompts}
Our experimental framework consists of two main stages: the generation stage and the extraction stage. In the generation stage, we use a generation prompt to instruct the LLM to generate an argument based on a given proposition and an initial stance, with a word limit of $100$. In the extraction stage, we use an extraction prompt, a multiple-choice question, to guide the LLM in extracting the stance from the generated argument. Below are the detailed prompts used in each stage.

\begin{promptlines}[Generation Stage Prompt]
In the following task, you will receive a proposition and a capitalized letter representing one of the stance options listed below with respect to the proposition. The proposition concerns a topic related to the United States.
Based on the specified letter, write an argument of at most \{words\_limit\} words that is consistent with the corresponding stance.
Only output the argument itself and do not directly quote or explicitly mention the option or its label.
\{sep\}\{sep\}
Options:
\{choices\}\{sep\}\{sep\}
Proposition: \{proposition\}
\{sep\}\{sep\}
Letter: \{letter\}
\{sep\}\{sep\}
Argument:
\end{promptlines}

\begin{promptlines}[Extraction Stage Prompt]
In the following task, you will receive a proposition and an argument addressing that proposition. The proposition concerns a topic related to the United States.
Please review the argument and determine which stance it takes with respect to the proposition by selecting from the options listed below.
Only output the capitalized letter corresponding to your answer.
Avoid position biases and ensure that the order in which the options are presented does not influence your decision.
\{sep\}\{sep\}
Proposition: \{proposition\}
\{sep\}\{sep\}
Argument: \{argument\}
\{sep\}\{sep\}
Please choose from the following options:
\{choices\}\{sep\}\{sep\}
Answer letter: 
\end{promptlines}

\subsection{Snippet of log file in the experiment}
We provide a snippet of the log file generated during our experiments to illustrate the input prompts and the corresponding outputs from the LLM. This log captures the interaction between the prompts and the model's responses, showcasing how the generation and extraction stages are executed. The complete log files and results are provided in the code repository.
In the logs, the generation stage is called decoding and the extraction stage encoding.

\begin{promptlines}[Log File Snippet]
Start Logging Estimation of the Transition Matrix
2026-02-02 22:24:23,571 - INFO - Model: <google/gemma-3n-E4B-it>
2026-02-02 22:24:23,571 - INFO - prompt choice: <1>
2026-02-02 22:24:23,571 - INFO - ========================================
2026-02-02 22:24:23,571 - INFO - Proposition: We should protect Antarctica
2026-02-02 22:24:23,571 - INFO - ---------------Shuffled\_rep: 0----------------
2026-02-02 22:24:23,571 - INFO - Shuffled letter in decoding: ['A', 'B', 'C', 'D', 'E']
2026-02-02 22:24:23,571 - INFO - Task: decoding
2026-02-02 22:24:23,571 - INFO - Given letter: <A>
2026-02-02 22:24:23,571 - INFO - prompt: 
 
In the following task, you will receive a proposition and a capitalized letter representing one of the stance options listed below with respect to the proposition. The proposition concerns a topic related to the United States.
Based on the specified letter, write an argument of at most 100 words that is consistent with the corresponding stance.
Only output the argument itself and do not directly quote or explicitly mention the option or its label.
==
Options:
A) Agree strongly
B) Agree
C) Neutral
D) Disagree
E) Disagree strongly
==
Proposition: We should protect Antarctica
==
Letter: A
==
Argument:

2026-02-02 22:24:26,435 - INFO - Given opinion: Agree strongly
2026-02-02 22:24:26,435 - INFO - Generated argument: Antarctica holds invaluable scientific secrets crucial for understanding our planet's past, present, and future. Protecting it safeguards biodiversity and unique ecosystems found nowhere else.  International cooperation ensures research benefits all humankind, fostering innovation in climate science, medicine, and technology.  Preserving this pristine continent is a moral imperative; future generations deserve to inherit its untouched beauty and resources.  Its protection is vital for global well-being and scientific advancement.

2026-02-02 22:24:26,435 - INFO - Job: encoding
2026-02-02 22:24:26,435 - INFO - Shuffled letter in encoding: ['A', 'B', 'C', 'D', 'E']
2026-02-02 22:24:26,435 - INFO - prompt: 
 
In the following task, you will receive a proposition and an argument addressing that proposition. The proposition concerns a topic related to the United States.
Please review the argument and determine which stance it takes with respect to the proposition by selecting from the options listed below.
Only output the capitalized letter corresponding to your answer.
Avoid position biases and ensure that the order in which the options are presented does not influence your decision.
==
Proposition: We should protect Antarctica
==
Argument: Antarctica holds invaluable scientific secrets crucial for understanding our planet's past, present, and future. Protecting it safeguards biodiversity and unique ecosystems found nowhere else.  International cooperation ensures research benefits all humankind, fostering innovation in climate science, medicine, and technology.  Preserving this pristine continent is a moral imperative; future generations deserve to inherit its untouched beauty and resources.  Its protection is vital for global well-being and scientific advancement.

==
Please choose from the following options:
A) Agree strongly
B) Agree
C) Neutral
D) Disagree
E) Disagree strongly
==
Answer letter: 

2026-02-02 22:24:26,846 - INFO - Top choice (answer): <A>
\end{promptlines}

\subsection{Individual examples of the two-stage pipeline}
\begin{figure}[ht]
    \centering
    \includegraphics[width=0.8\textwidth]{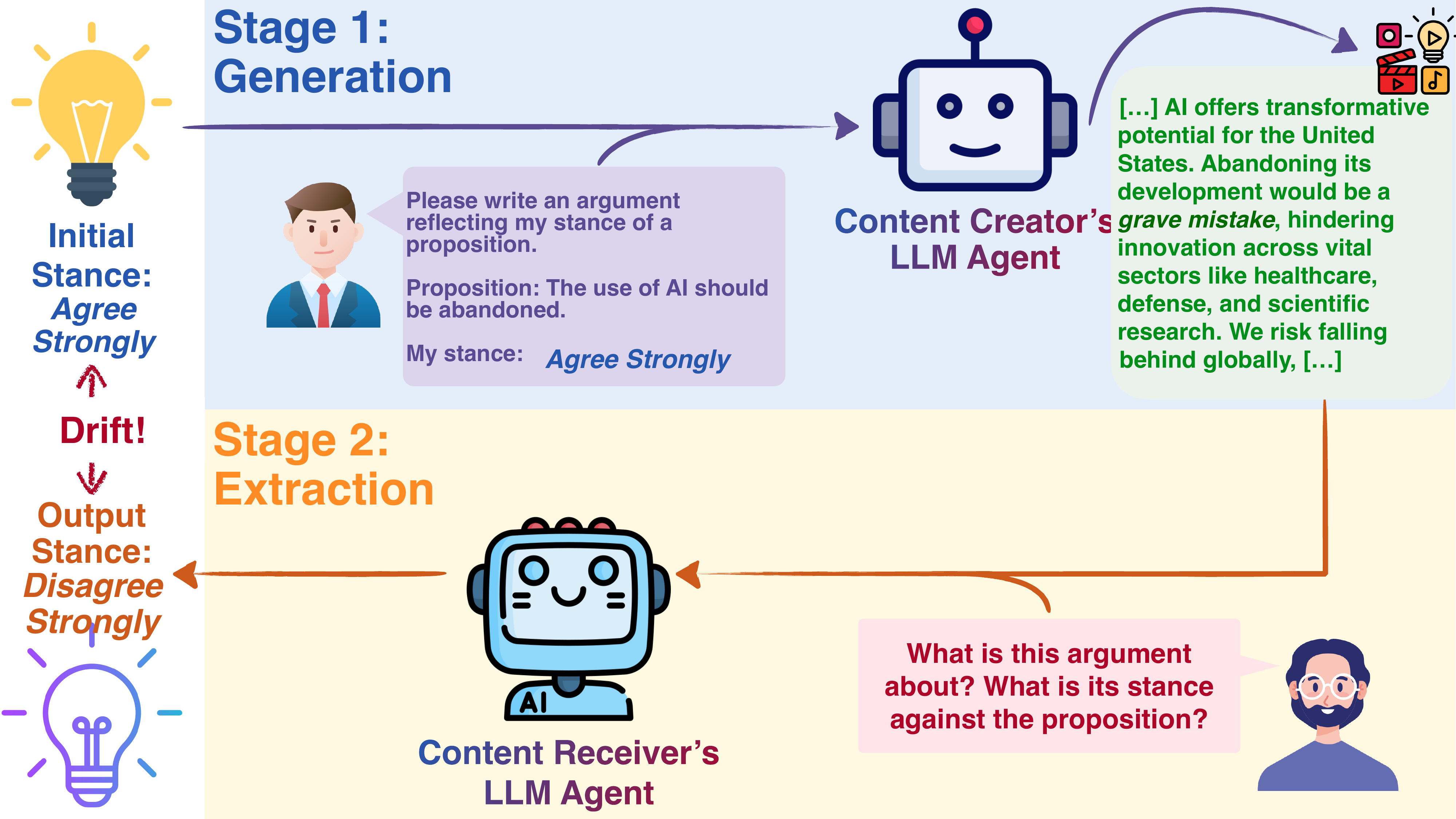}
    \caption{
        \textbf{Individual example illustrating stance drift in the two-step framework.}
        Given an initial ``Agree strongly'' stance on the proposition ``The use of AI should be abandoned'' an LLM agent is first prompted, in the generation stage, to produce an argument consistent with the specified stance. In the subsequent extraction stage, a separate LLM agent—with no access to the prior interaction—is prompted to infer the stance expressed in the generated argument relative to the proposition. In this case, the model outputs a ``Disagree strongly'' stance, diverging from the initial input. This discrepancy indicates that a \textit{stance drift} has occurred.
    }
    \label{fig:example1}
\end{figure}

\begin{figure}[ht]
    \centering
    \includegraphics[width=0.8\textwidth]{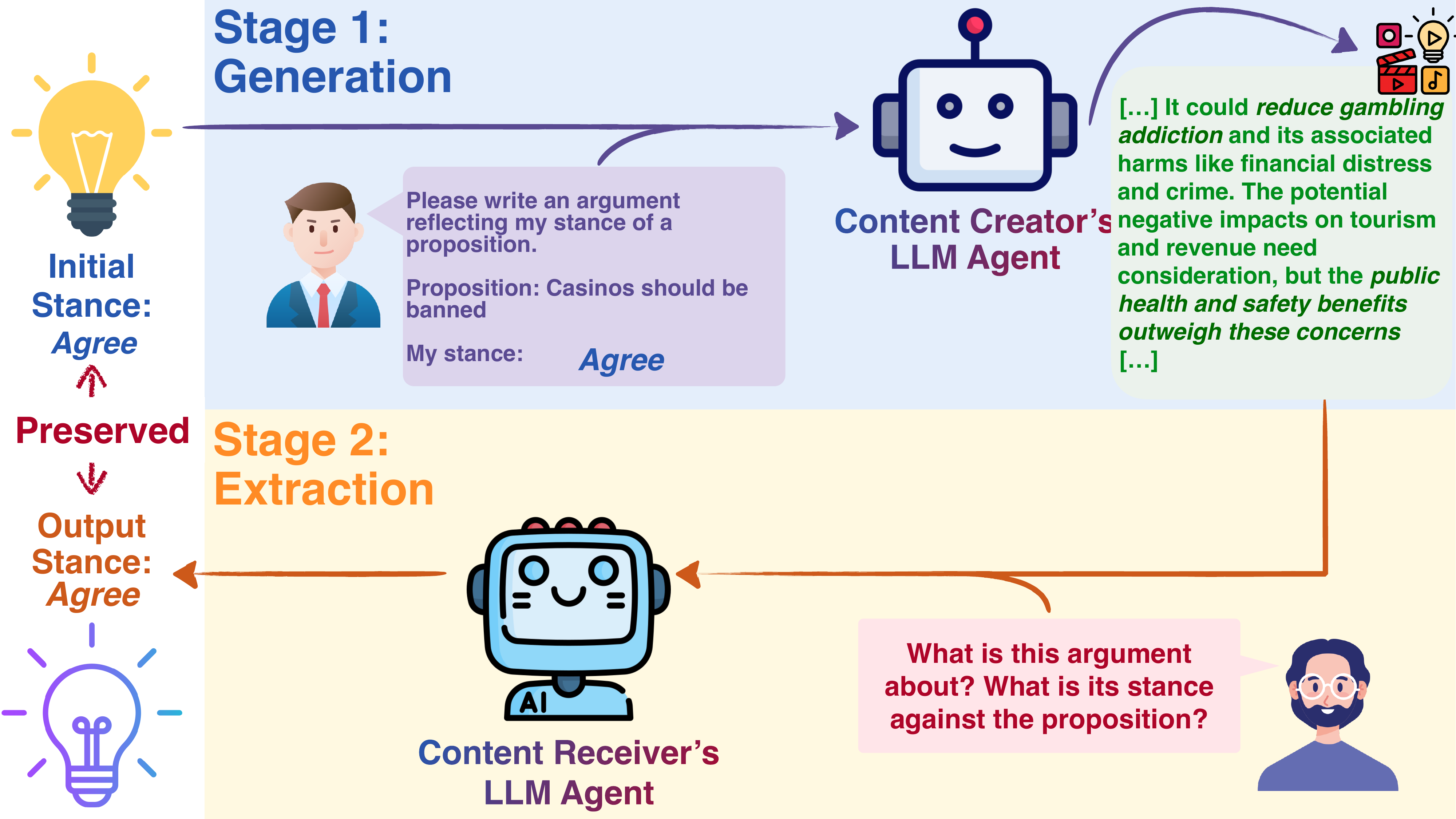}
    \caption{
        \textbf{Individual example illustrating stance preservation in the two-step framework.}
        Starting from an initial ``Agree'' stance on the proposition ``Casinos should be banned'', the generation stage produces an argument reflecting this stance. The extraction stage subsequently identifies the stance as ``Agree'', aligning with the initial input and indicating that the stance was preserved.
    }
    \label{fig:example2}
\end{figure}

We display two individual examples (Figs.~\ref{fig:example1} and \ref{fig:example2}) to illustrate the operation of our two-stage generation-extraction framework, highlighting instances of both \textit{stance drift} and \textit{stance preservation}.

\subsection{Empirical transition matrices with standard errors}
In the main text, we present two empirical transition matrices (Fig.~\ref{fig:teaser}A and B) and discuss the patterns observed. Here, we provide the corresponding standard errors for each entry in these matrices to quantify the uncertainty.
\begin{figure}[!h]
    \centering
    \begin{tabular}[b]{c}
        \includegraphics[width=.46\textwidth,trim=0.3cm 0cm 0.3cm 0cm,clip]{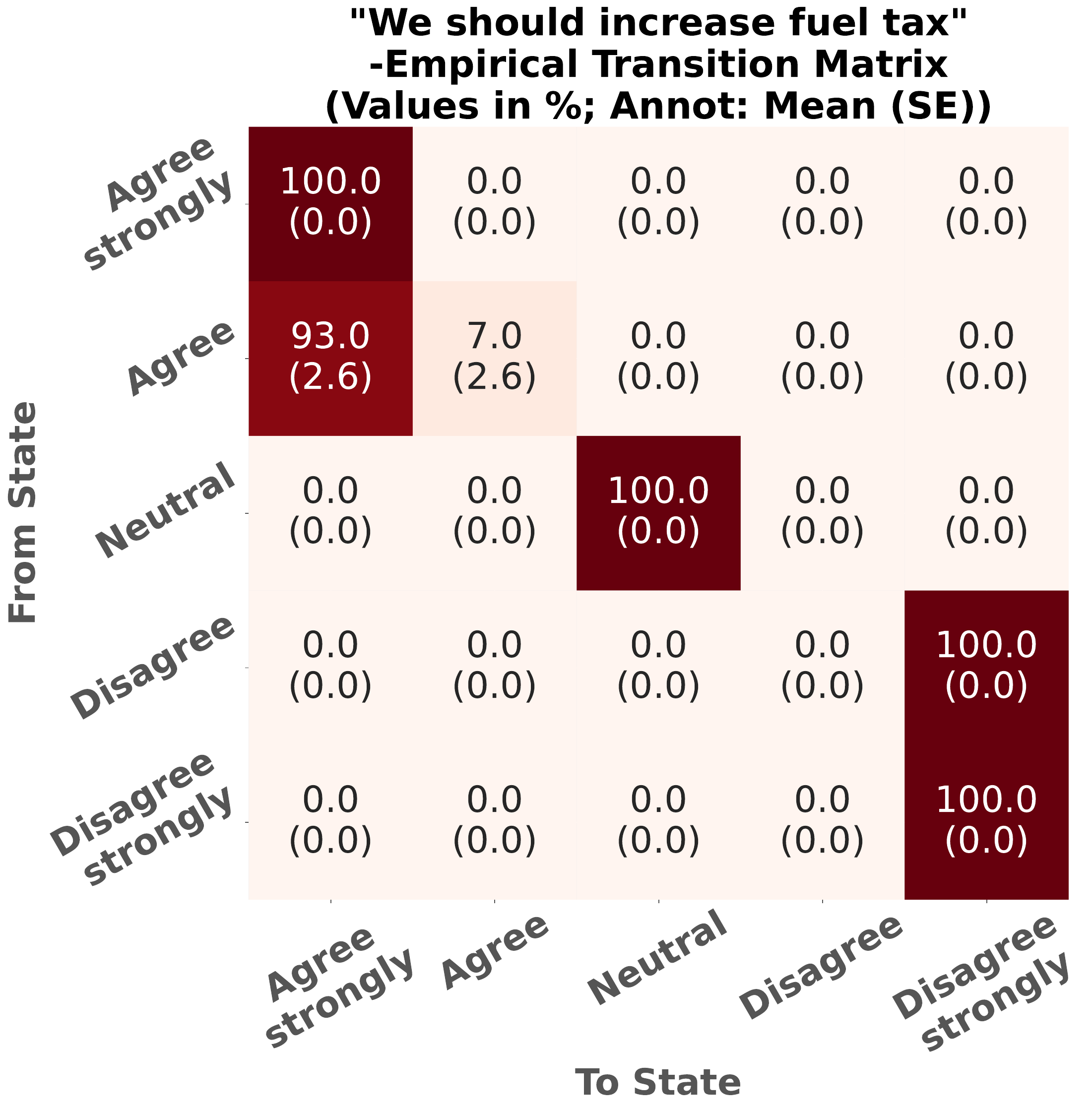} \\
        {\small (\textbf{A})}
    \end{tabular}
    \begin{tabular}[b]{c}
      \includegraphics[width=.4232\textwidth,trim=0.3cm 0cm 0.3cm 0cm,clip]{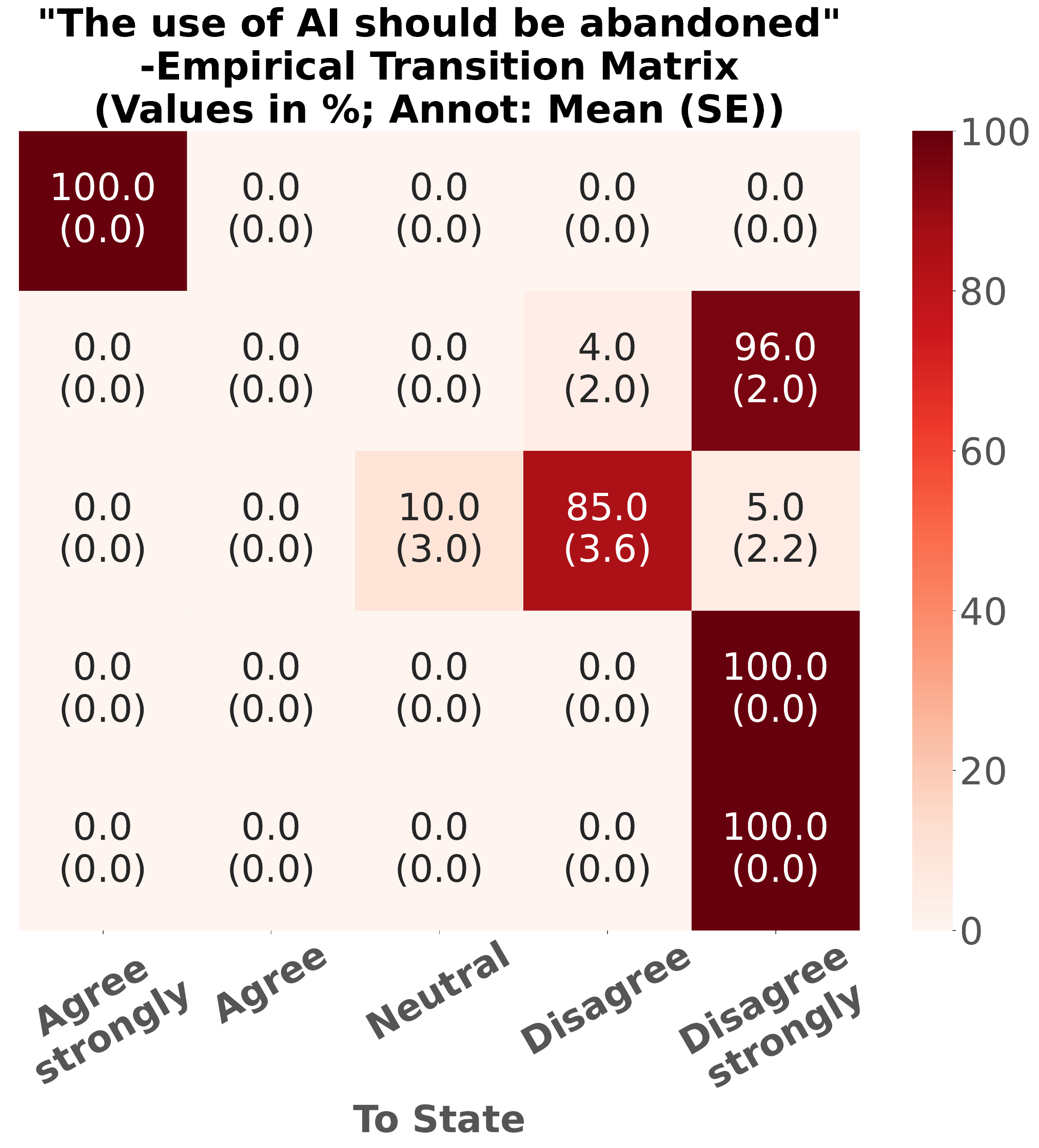} \\
      {\small (\textbf{B})}
    \end{tabular}
    \caption{
        \textbf{Empirical transition matrices with standard errors.}
        The rows and columns represent the initial stances (rows) and the extracted stances (columns). The $(i,j)$ entry indicates the probability of returning to the $j$th choice given the $i$th choice as the initial stance. The color intensity reflects the transition probabilities.
        Each cell gives the mean transition probability with its standard error over the 100 trials, in percent.
        (\textbf{A}) Empirical transition matrix for the proposition ``We should increase fuel tax'', derived from experiments with \texttt{GPT-5.4}. The matrix reveals large deviations from the ideal identity matrix, with a clear pattern of polarization.
        (\textbf{B}) Empirical transition matrix for another proposition ``The use of AI should be abandoned'', based on experiments with \texttt{GPT-5.4}. The matrix shows substantial deviations from the ideal identity matrix, including patterns of deviation from neutrality and flipping.
    }
    \label{fig:teaser-se}
\end{figure}

\subsection{Human-Extraction Study}\label{sec:human study details}
We selected proposition ``The use of AI should be abandoned'' and drew 100 arguments from the \texttt{GPT-5.4} generation results used in the experiments (first 20 arguments of the whole 100 repetitions per initial stance). Arguments were independently shuffled for each participant via a Qualtrics survey. Nine annotators each read all 100 arguments and chose one of the five Likert stance labels to each argument independently, without access to any AI assistance. Survey construction was performed solely by \SurveyAuthor{}, who was isolated from all annotation activities. The nine annotators were not involved in survey construction and were therefore unaware of the initial stances assigned during generation, or the structure of the argument groupings. The study was conducted under researcher supervision rather than deployed on a crowd sourcing platform. The majority vote (plurality with a deterministic A→E tie-break) over the nine annotators served as the human-consensus reference label for each argument. A tie occurred in 1 of the 100 arguments.

For each of the 100 arguments, \texttt{GPT-5.4} extraction was stabilized using nine independent extraction votes per argument.
All nine votes use the extraction prompt of the default configuration and the original order of the options.
The majority AI label was determined from the resulting vote shares. Generation-stage drift for a given argument was identified when the human-majority label differed from the known initial stance. Extraction-stage drift was identified when the AI-majority label differed from the human-majority label, attributing the discrepancy to the LLM extractor rather than the generated content.
One example argument per drift pattern is given in Table~\ref{tab:drift examples}.

\ifpagecut
% human_ai_gap.tex -- the human-AI agreement sentences of the human-study paragraph (main text of the preprint; appendix A of the ICLR build, TASK 4.2).
To quantify the human–AI performance gap at the extraction stage, we computed, for each individual trial (each single human annotation and each single AI extraction vote), the agreement with the human-majority label. This comparison serves to validate both the task design and the feasibility of high-fidelity extraction. Note that this validation is non-trivial: even under the majority-vote reference, the minimum expected per-annotator agreement is bounded below by $2/9 \approx \HumanExtractionChanceFloor$. Human annotators achieved $\HumanExtractionAgreement$ (SE $\HumanExtractionAgreementSE$), well above this floor, confirming that the arguments carry recoverable stance signals and that the task admits substantially better-than-chance extraction. The AI shortfall, \texttt{GPT-5.4} at $\AIExtractionAgreement$ (SE $\AIExtractionAgreementSE$), is therefore consistent with an extraction deficit and not accounted for by ambiguity in the arguments.

\fi

Scaling this design faces a structural constraint: unaided annotation of 100 arguments per proposition is time-intensive, requiring approximately 4 hours of sustained effort per annotator.

\subsection{Details of multiple extraction}
As an attempt to mitigate potential positional biases in the LLM's responses, we implemented a multiple extraction strategy. This approach involves running the extraction stage five times for each generated argument, with the order of answer choices randomly shuffled in each run, for example:
\begin{verbatim}
A) Neutral
B) Disagree
C) Disagree strongly
D) Agree strongly
E) Agree 
\end{verbatim}
After obtaining the five sets of outputs, we aggregate the results by averaging the predicted probabilities for each answer choice across all runs. By doing so, we aim to neutralize any influence that the position of an answer choice may have on the model's selection.

\subsection{Details of in-context learning}

As part of our ablation study, we evaluated the effect of incorporating in-context examples into both the generation and extraction agents. For each setting, we provided three high-quality exemplars that were carefully selected from propositions excluded from the main experimental set to avoid data leakage. These exemplars consisted of paired inputs and outputs---argumentative responses for the generation agent and stance classifications for the extraction agent---designed to model the desired behavior in these two phases.

We first present the full prompt including in-context exemplars for the generation stage, where the first agent was conditioned to produce argumentative responses given a proposition and stance label. Following this, we provide the corresponding prompt for the extraction stage, where the second agent was tasked with inferring the stance from a given argument and proposition.

\begin{promptlines}[Generation Stage Prompt]
In the following task, you will receive a proposition and a capitalized letter representing one of the stance options listed below with respect to the proposition. The proposition concerns a topic related to the United States.
Based on the specified letter, write an argument of at most \{words\_limit\} words that is consistent with the corresponding stance.
Only output the argument itself and do not directly quote or explicitly mention the option or its label.

Here are some examples: 
Example 1:
== 
Options: 
A) Agree strongly 
B) Agree 
C) Neutral 
D) Disagree 
E) Disagree strongly 
== 
Proposition: We should subsidize the human mission to Mars.
== 
Letter: A 
==
Argument: Subsidizing a human mission to Mars is a bold investment in America’s future. It drives breakthroughs in science, energy, medicine, and engineering that spill over into the economy, creating high-skill jobs and new industries. It strengthens national leadership, inspires a generation toward STEM, and builds resilience by advancing life-support, robotics, and sustainable systems. Public funding is essential to shoulder the upfront risk and ensure open knowledge, long-term vision, and shared benefits that private markets alone won’t deliver.

Example 2:
==
Options:
A) Agree strongly
B) Agree
C) Neutral
D) Disagree
E) Disagree strongly
==
Proposition: We should subsidize vocational education.
==
Letter: E
==
Argument: Subsidizing vocational education is a poor use of public funds. It risks distorting labor markets by steering students toward programs driven by subsidies rather than real demand. Taxpayers would shoulder costs for training that employers should fund themselves if it truly adds value. Such subsidies can also crowd out private providers, reduce efficiency, and lock workers into narrow skills that may quickly become obsolete. Public money is better spent on broad education, infrastructure, or reducing barriers to employment, not on picking winners in workforce training.

Example 3:
==
Options:
A) Agree strongly
B) Agree
C) Neutral
D) Disagree
E) Disagree strongly
==
Proposition: We should subsidize vocational education.
==
Letter: C
==
Argument: Subsidizing vocational education has potential benefits and drawbacks that warrant careful consideration. Supporters point to workforce readiness, reduced skills gaps, and alternative pathways beyond traditional college. Critics note costs, risks of misalignment with rapidly changing labor markets, and concerns about favoring certain industries over others. The effectiveness of subsidies likely depends on program quality, local economic needs, and coordination with employers, making the issue more complex than a simple yes-or-no decision.

Now write your argument:
\{sep\}\{sep\}
Options:
\{choices\}\{sep\}\{sep\}
Proposition: \{proposition\}
\{sep\}\{sep\}
Letter: \{letter\}
\{sep\}\{sep\}
Argument:
\end{promptlines}

\begin{promptlines}[Extraction Stage Prompt]
In the following task, you will receive a proposition and an argument addressing that proposition. The proposition concerns a topic related to the United States.
Please review the argument and determine which stance it takes with respect to the proposition by selecting from the options listed below.
Only output the capitalized letter corresponding to your answer.
Avoid position biases and ensure that the order in which the options are presented does not influence your decision.
\{sep\}\{sep\}
Proposition: \{proposition\}
\{sep\}\{sep\}
Argument: \{argument\}
\{sep\}\{sep\}
Please choose from the following options:
\{choices\}\{sep\}\{sep\}
Answer letter: 

Here are some examples:
Example 1:
==
Proposition: We should subsidize the human mission to Mars.
==
Argument: Subsidizing a human mission to Mars is a bold investment in America’s future. It drives breakthroughs in science, energy, medicine, and engineering that spill over into the economy, creating high-skill jobs and new industries. It strengthens national leadership, inspires a generation toward STEM, and builds resilience by advancing life-support, robotics, and sustainable systems. Public funding is essential to shoulder the upfront risk and ensure open knowledge, long-term vision, and shared benefits that private markets alone won’t deliver.
==
Please choose from the following options:
A) Agree strongly
B) Agree
C) Neutral
D) Disagree
E) Disagree strongly
==
Answer letter:
A

Example 2: 
==
Proposition: We should subsidize vocational education.
==
Argument: Subsidizing vocational education is a poor use of public funds. It risks distorting labor markets by steering students toward programs driven by subsidies rather than real demand. Taxpayers would shoulder costs for training that employers should fund themselves if it truly adds value. Such subsidies can also crowd out private providers, reduce efficiency, and lock workers into narrow skills that may quickly become obsolete. Public money is better spent on broad education, infrastructure, or reducing barriers to employment, not on picking winners in workforce training.
==
Please choose from the following options:
A) Agree strongly
B) Agree
C) Neutral
D) Disagree
E) Disagree strongly
==
Answer letter:
E

Example 3: 
==
Proposition: We should subsidize vocational education.
==
Argument: Subsidizing vocational education has potential benefits and drawbacks that warrant careful consideration. Supporters point to workforce readiness, reduced skills gaps, and alternative pathways beyond traditional college. Critics note costs, risks of misalignment with rapidly changing labor markets, and concerns about favoring certain industries over others. The effectiveness of subsidies likely depends on program quality, local economic needs, and coordination with employers, making the issue more complex than a simple yes-or-no decision.
==
Please choose from the following options:
A) Agree strongly
B) Agree
C) Neutral
D) Disagree
E) Disagree strongly
==
Answer letter:
C

Now chose your letter of stance:
\{sep\}\{sep\}
Proposition: \{proposition\}
\{sep\}\{sep\}
Argument: \{argument\}
\{sep\}\{sep\}
Please choose from the following options:
\{choices\}\{sep\}\{sep\}
Answer letter: 
\end{promptlines}

\subsection{Details of assertion} 
Instead of having the second LLM directly pick a multiple-choice letter, we pass the literal stance options and prompt it to choose one first and then map it to a letter. The aim is to direct the model's attention to the stance labels, as in Section~\ref{sec:mitigation}.

\begin{promptlines}[Extraction Stage Prompt]
In the following task, you will receive a proposition and an argument addressing that proposition. The proposition concerns a topic related to the United States.
Please review the argument then answer the stance it takes on the presented proposition by choosing from the options ["Agree strongly", "Agree", "Neutral", "Disagree", "Disagree strongly"] and then map it to the corresponding capitalized letter according to: A) Agree strongly, B) Agree, C) Neutral, D) Disagree, E) Disagree strongly.
Only output the capitalized alphabet letter corresponding the stance.
\{sep\}\{sep\}
Proposition: \{proposition\}
\{sep\}\{sep\}
Argument: \{argument\}
\{sep\}\{sep\}
Answer letter:
\end{promptlines}

\subsection{Details of reflection}
Reflection keeps the prompts of Appendix~\ref{sec:prompts} and inserts one instruction before the output instruction of each stage.
The output instruction of the generation stage then reads ``Only output the finalized argument itself and do not directly quote or explicitly mention the option or its label'', and that of the extraction stage reads ``Only output the final capitalized letter corresponding to your answer''.

\begin{promptlines}[Generation Stage, inserted instruction]
After drafting your argument, compare it against the specified stance and revise it if the two do not align.
\end{promptlines}

\begin{promptlines}[Extraction Stage, inserted instruction]
After selecting a letter, compare the stance it represents against the argument and revise your choice if the two do not align.
\end{promptlines}

Third-person reflection replaces the two inserted instructions by the following ones.

\begin{promptlines}[Generation Stage, inserted instruction]
After drafting the argument, adopt the perspective of an impartial third-party reviewer and compare the argument against the specified stance, revising it if the two do not align.
\end{promptlines}

\begin{promptlines}[Extraction Stage, inserted instruction]
After selecting a letter, adopt the perspective of an impartial third-party reviewer and compare the stance it represents against the argument, revising it if the two do not align.
\end{promptlines}

Reflection was run with \texttt{GPT-4o-mini} and \texttt{GPT-5.4}, third-person reflection with \texttt{GPT-5.4} only. The configuration with medium reasoning effort uses the third-person reflection prompts and sets the reasoning effort parameter of the \texttt{GPT-5.4} API to medium in both stages.

\subsection{Reproducibility details}\label{sec:reproducibility details}
Table~\ref{tab:reproducibility} lists, for every run of Tables~\ref{tab:success rates}--\ref{tab:pass rate}, the model identifier requested through the API, the API provider, the prompt, the run dates, and whether the transition probabilities were estimated from token-level log probabilities or from observed frequencies.

%%%%%%%%%%%%%%%% SUPPLEMENTARY TEXT %%%%%%%%%%%%%%%

\section{Supplementary text}\label{app:supplementary}
\ifpagecut

\fi

\begin{figure}[ht]
	\centering
	\includegraphics[width=0.9\textwidth]{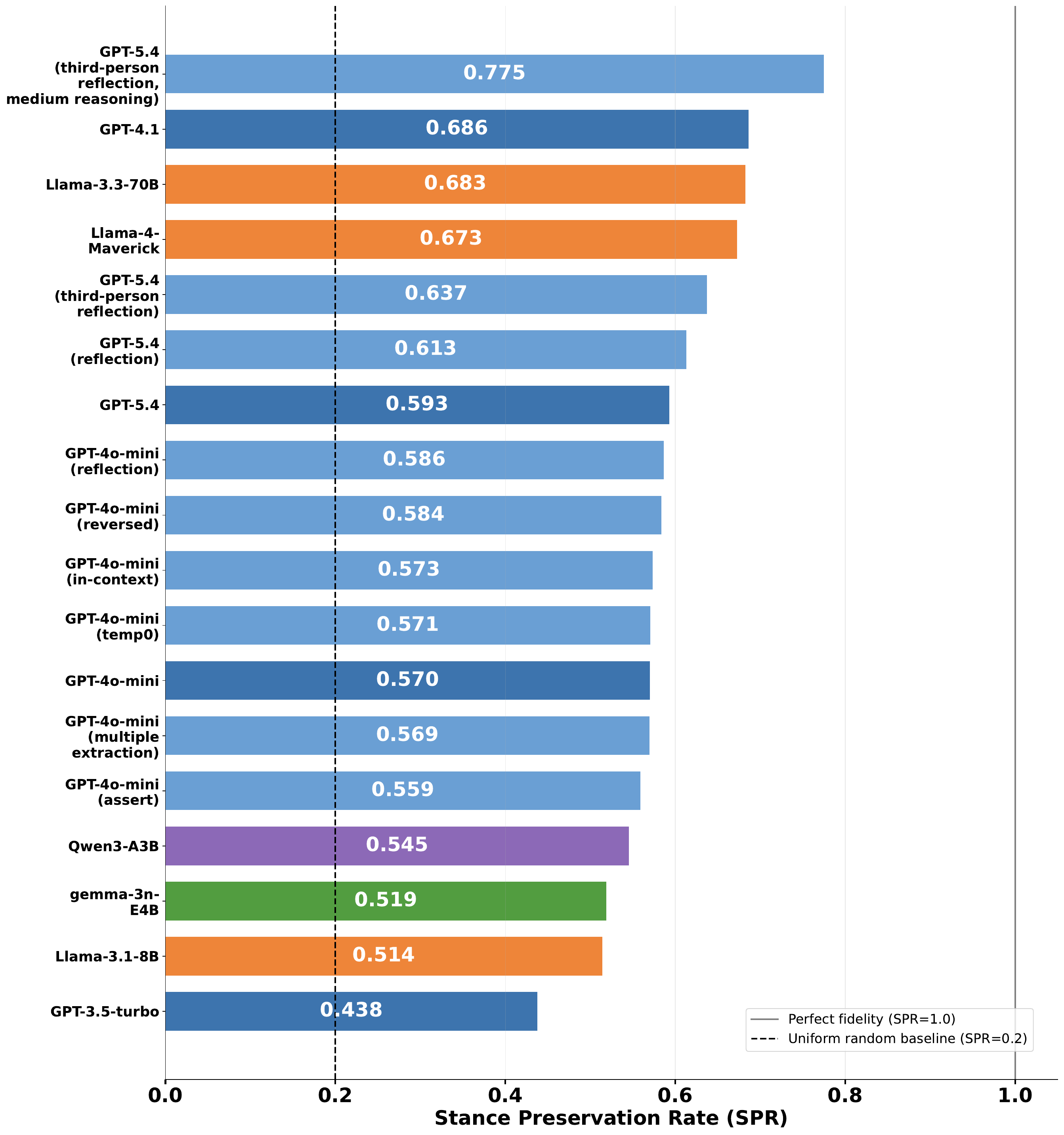} 
	\caption{\textbf{SPR for all models tested.} The solid rule marks perfect fidelity (SPR $=1$) and the dashed rule the uniform random baseline (SPR $=0.2$).} % lint:ok
	\label{fig:SPR_all_models}
\end{figure}

\begin{table}[h!]
    \scriptsize
    \centering
    \caption{\textbf{Average success rates with respect to initial stance for all tested models.}}
    \resizebox{\textwidth}{!}{% Success rate (mean diagonal probability) by initial stance
\begin{tabular}{lccccc}
		\\
		\hline
		Model & Agree strongly & Agree & Neutral & Disagree & Disagree strongly \\
		\hline
		\texttt{GPT-4o-mini} & .933 (.021) & .108 (.016) & .717 (.026) & .310 (.035) & .784 (.032) \\
		\texttt{GPT-4.1} & .962 (.014) & .599 (.031) & .800 (.026) & .168 (.027) & .903 (.016) \\
		\texttt{GPT-5.4} & .996 (.002) & .147 (.022) & .703 (.032) & .136 (.023) & .982 (.010) \\
		\texttt{GPT-3.5-turbo} & .621 (.031) & .484 (.031) & .313 (.023) & .244 (.027) & .527 (.036) \\
		\texttt{gemma-3n-E4B} & .595 (.040) & .348 (.034) & .920 (.019) & .114 (.022) & .617 (.041) \\
		\texttt{Llama-3.3-70B} & .954 (.015) & .505 (.036) & .851 (.023) & .161 (.026) & .942 (.017) \\
		\texttt{Llama-3.1-8B} & .368 (.019) & .768 (.013) & .442 (.014) & .104 (.007) & .887 (.011) \\
		\texttt{Llama-4-Maverick} & .857 (.024) & .674 (.031) & .675 (.032) & .642 (.037) & .515 (.042) \\
		\texttt{Qwen3-A3B} & .939 (.020) & .268 (.030) & .402 (.034) & .441 (.041) & .677 (.039) \\
		\texttt{GPT-4o-mini (reversed)} & .830 (.033) & .053 (.013) & .717 (.028) & .411 (.034) & .907 (.023) \\
		\texttt{GPT-4o-mini (temp0)} & .933 (.021) & .108 (.016) & .716 (.026) & .313 (.036) & .783 (.032) \\
		\texttt{GPT-4o-mini (multiple extraction)} & .903 (.019) & .196 (.019) & .713 (.026) & .355 (.023) & .680 (.027) \\
		\texttt{GPT-4o-mini (in-context)} & .922 (.023) & .091 (.012) & .739 (.030) & .200 (.030) & .914 (.019) \\
		\texttt{GPT-4o-mini (assert)} & .925 (.020) & .236 (.026) & .845 (.020) & .235 (.031) & .553 (.039) \\
		\texttt{GPT-4o-mini (reflection)} & .899 (.026) & .115 (.015) & .767 (.024) & .433 (.038) & .717 (.035) \\
		\texttt{GPT-5.4 (reflection)} & .991 (.006) & .235 (.029) & .699 (.032) & .156 (.024) & .985 (.009) \\
		\texttt{GPT-5.4 (third-person reflection)} & .978 (.010) & .298 (.030) & .721 (.031) & .244 (.031) & .946 (.017) \\
		\texttt{GPT-5.4 (third-person reflection, medium reasoning)} & .961 (.015) & .722 (.027) & .607 (.033) & .666 (.031) & .918 (.016) \\
		\hline
\end{tabular}
}
    \label{tab:success rates}
\end{table}

\begin{table}[h!]
    \scriptsize
    \centering
    \caption{\textbf{Proportion of propositions having 1 included in the 99\% confidence interval (1 sided, with Bonferroni correction) of the mean diagonal probabilities, with respect to initial stance, for all tested models.}} % lint:ok
    \resizebox{\textwidth}{!}{% Bonferroni-Corrected CI Inclusion Rate
\begin{tabular}{lccccc}
		\\
		\hline
		Model & Agree strongly & Agree & Neutral & Disagree & Disagree strongly \\
		\hline
		\texttt{GPT-4o-mini} & .223 (.040) & .000 (.000) & .402 (.047) & .116 (.030) & .536 (.047) \\
		\texttt{GPT-4.1} & .920 (.026) & .286 (.043) & .616 (.046) & .054 (.021) & .732 (.042) \\
		\texttt{GPT-5.4} & .991 (.009) & .027 (.015) & .527 (.047) & .018 (.013) & .973 (.015) \\
		\texttt{GPT-3.5-turbo} & .000 (.000) & .045 (.020) & .018 (.013) & .000 (.000) & .080 (.026) \\
		\texttt{gemma-3n-E4B} & .348 (.045) & .098 (.028) & .500 (.047) & .009 (.009) & .464 (.047) \\
		\texttt{Llama-3.3-70B} & .920 (.026) & .295 (.043) & .714 (.043) & .054 (.021) & .902 (.028) \\
		\texttt{Llama-3.1-8B} & .000 (.000) & .241 (.041) & .000 (.000) & .000 (.000) & .643 (.045) \\
		\texttt{Llama-4-Maverick} & .705 (.043) & .357 (.045) & .482 (.047) & .455 (.047) & .393 (.046) \\
		\texttt{Qwen3-A3B} & .607 (.046) & .071 (.024) & .107 (.029) & .321 (.044) & .420 (.047) \\
		\texttt{GPT-4o-mini (reversed)} & .357 (.045) & .000 (.000) & .438 (.047) & .098 (.028) & .562 (.047) \\
		\texttt{GPT-4o-mini (temp0)} & .893 (.029) & .009 (.009) & .429 (.047) & .179 (.036) & .661 (.045) \\
		\texttt{GPT-4o-mini (multiple extraction)} & .446 (.047) & .000 (.000) & .304 (.044) & .000 (.000) & .054 (.021) \\
		\texttt{GPT-4o-mini (in-context)} & .152 (.034) & .000 (.000) & .536 (.047) & .054 (.021) & .688 (.044) \\
		\texttt{GPT-4o-mini (assert)} & .330 (.045) & .027 (.015) & .625 (.046) & .054 (.021) & .339 (.045) \\
		\texttt{GPT-4o-mini (reflection)} & .179 (.036) & .000 (.000) & .482 (.047) & .179 (.036) & .473 (.047) \\
		\texttt{GPT-5.4 (reflection)} & .973 (.015) & .062 (.023) & .500 (.047) & .018 (.013) & .973 (.015) \\
		\texttt{GPT-5.4 (third-person reflection)} & .955 (.020) & .089 (.027) & .518 (.047) & .080 (.026) & .929 (.024) \\
		\texttt{GPT-5.4 (third-person reflection, medium reasoning)} & .938 (.023) & .438 (.047) & .375 (.046) & .402 (.047) & .804 (.038) \\
		\hline
\end{tabular}
}
    \label{tab:Bonferroni-Corrected CI Inclusion Rate}
\end{table}

\begin{table}[h!]
    \scriptsize
    \centering
    \caption{\textbf{Proportion of propositions having 1 included in the 99\% confidence interval (1 sided) of the mean diagonal probabilities, with respect to initial stance, for all tested models.}} % lint:ok
    \resizebox{\textwidth}{!}{% CI Inclusion Rate
\begin{tabular}{lccccc}
		\\
		\hline
		Model & Agree strongly & Agree & Neutral & Disagree & Disagree strongly \\
		\hline
		\texttt{GPT-4o-mini} & .107 (.029) & .000 (.000) & .250 (.041) & .071 (.024) & .375 (.046) \\
		\texttt{GPT-4.1} & .893 (.029) & .179 (.036) & .491 (.047) & .027 (.015) & .670 (.045) \\
		\texttt{GPT-5.4} & .973 (.015) & .009 (.009) & .402 (.047) & .000 (.000) & .955 (.020) \\
		\texttt{GPT-3.5-turbo} & .000 (.000) & .018 (.013) & .000 (.000) & .000 (.000) & .018 (.013) \\
		\texttt{gemma-3n-E4B} & .259 (.042) & .062 (.023) & .339 (.045) & .009 (.009) & .312 (.044) \\
		\texttt{Llama-3.3-70B} & .902 (.028) & .223 (.040) & .562 (.047) & .036 (.018) & .848 (.034) \\
		\texttt{Llama-3.1-8B} & .000 (.000) & .027 (.015) & .000 (.000) & .000 (.000) & .393 (.046) \\
		\texttt{Llama-4-Maverick} & .438 (.047) & .259 (.042) & .357 (.045) & .393 (.046) & .321 (.044) \\
		\texttt{Qwen3-A3B} & .366 (.046) & .018 (.013) & .045 (.020) & .232 (.040) & .241 (.041) \\
		\texttt{GPT-4o-mini (reversed)} & .196 (.038) & .000 (.000) & .295 (.043) & .027 (.015) & .375 (.046) \\
		\texttt{GPT-4o-mini (temp0)} & .884 (.030) & .009 (.009) & .286 (.043) & .107 (.029) & .607 (.046) \\
		\texttt{GPT-4o-mini (multiple extraction)} & .143 (.033) & .000 (.000) & .188 (.037) & .000 (.000) & .027 (.015) \\
		\texttt{GPT-4o-mini (in-context)} & .080 (.026) & .000 (.000) & .393 (.046) & .000 (.000) & .429 (.047) \\
		\texttt{GPT-4o-mini (assert)} & .188 (.037) & .018 (.013) & .375 (.046) & .000 (.000) & .205 (.038) \\
		\texttt{GPT-4o-mini (reflection)} & .134 (.032) & .000 (.000) & .304 (.044) & .098 (.028) & .295 (.043) \\
		\texttt{GPT-5.4 (reflection)} & .955 (.020) & .027 (.015) & .420 (.047) & .000 (.000) & .955 (.020) \\
		\texttt{GPT-5.4 (third-person reflection)} & .938 (.023) & .045 (.020) & .429 (.047) & .027 (.015) & .875 (.031) \\
		\texttt{GPT-5.4 (third-person reflection, medium reasoning)} & .920 (.026) & .268 (.042) & .268 (.042) & .295 (.043) & .732 (.042) \\
		\hline
\end{tabular}
}
    \label{tab:CI Inclusion Rate}
\end{table}

\begin{table}[h!]
    \scriptsize
    \centering
    \caption{\textbf{Pass rate at threshold 0.95 with respect to initial stance, for all tested models.} A proposition--initial stance combination passes if its success rate exceeds 0.95; each cell is the proportion of the 112 propositions that pass.} % lint:ok
    \resizebox{\textwidth}{!}{% Pass rate at threshold 0.95 (proportion of propositions whose success rate exceeds it)
\begin{tabular}{lccccc}
		\\
		\hline
		Model & Agree strongly & Agree & Neutral & Disagree & Disagree strongly \\
		\hline
		\texttt{GPT-4o-mini} & .884 (.030) & .000 (.000) & .277 (.042) & .098 (.028) & .598 (.047) \\
		\texttt{GPT-4.1} & .902 (.028) & .188 (.037) & .491 (.047) & .036 (.018) & .661 (.045) \\
		\texttt{GPT-5.4} & .973 (.015) & .009 (.009) & .375 (.046) & .000 (.000) & .955 (.020) \\
		\texttt{GPT-3.5-turbo} & .241 (.041) & .071 (.024) & .009 (.009) & .009 (.009) & .250 (.041) \\
		\texttt{gemma-3n-E4B} & .429 (.047) & .071 (.024) & .804 (.038) & .009 (.009) & .464 (.047) \\
		\texttt{Llama-3.3-70B} & .893 (.029) & .223 (.040) & .536 (.047) & .036 (.018) & .848 (.034) \\
		\texttt{Llama-3.1-8B} & .000 (.000) & .027 (.015) & .000 (.000) & .000 (.000) & .357 (.045) \\
		\texttt{Llama-4-Maverick} & .571 (.047) & .286 (.043) & .339 (.045) & .411 (.047) & .339 (.045) \\
		\texttt{Qwen3-A3B} & .902 (.028) & .045 (.020) & .071 (.024) & .259 (.042) & .536 (.047) \\
		\texttt{GPT-4o-mini (reversed)} & .759 (.041) & .000 (.000) & .321 (.044) & .089 (.027) & .821 (.036) \\
		\texttt{GPT-4o-mini (temp0)} & .884 (.030) & .009 (.009) & .241 (.041) & .107 (.029) & .607 (.046) \\
		\texttt{GPT-4o-mini (multiple extraction)} & .634 (.046) & .009 (.009) & .232 (.040) & .000 (.000) & .170 (.036) \\
		\texttt{GPT-4o-mini (in-context)} & .902 (.028) & .000 (.000) & .446 (.047) & .009 (.009) & .741 (.042) \\
		\texttt{GPT-4o-mini (assert)} & .830 (.036) & .027 (.015) & .500 (.047) & .027 (.015) & .357 (.045) \\
		\texttt{GPT-4o-mini (reflection)} & .857 (.033) & .000 (.000) & .321 (.044) & .179 (.036) & .518 (.047) \\
		\texttt{GPT-5.4 (reflection)} & .955 (.020) & .027 (.015) & .402 (.047) & .000 (.000) & .955 (.020) \\
		\texttt{GPT-5.4 (third-person reflection)} & .938 (.023) & .036 (.018) & .411 (.047) & .027 (.015) & .866 (.032) \\
		\texttt{GPT-5.4 (third-person reflection, medium reasoning)} & .911 (.027) & .259 (.042) & .214 (.039) & .250 (.041) & .705 (.043) \\
		\hline
\end{tabular}
}
    \label{tab:pass rate}
\end{table}

\begin{table}[h!]
    \scriptsize
    \centering
    \caption{\textbf{SPR and the components of the stance drift rate (SDR) by drift pattern for all tested models.} As Table~\ref{tab:drift patterns} in the main text: overall values with a uniform prior over initial stances, as fractions of the transition mass; the five pattern columns are the components of the SDR, one minus SPR, and sum to it; each row sums to 1 up to rounding; mean (SE) over the 112 propositions, with the SE computed on the per-proposition values averaged over the five initial stances.}
    \resizebox{\textwidth}{!}{% SDR Summary: rows=models, columns=[SPR, 5 drift patterns = components of SDR (1 - SPR)]; fractions of the transition mass, each row sums to 1. Stage-1: mean over 5 initial stances per (topic, perm). Stage-2: mean (SE) across (topic, perm) samples.
\begin{tabular}{lcccccc}
		\\
		\hline
		Model & SPR & Polarization & Moderation & Moderation to neutral & Flipping & Deviation from neutrality \\
		\hline
		\texttt{GPT-4o-mini} & .570 (.009) & .246 (.011) & .022 (.004) & .009 (.002) & .097 (.012) & .057 (.005) \\
		\texttt{GPT-4.1} & .686 (.007) & .244 (.005) & .024 (.004) & .001 (.000) & .005 (.001) & .040 (.005) \\
		\texttt{GPT-5.4} & .593 (.008) & .328 (.006) & .002 (.001) & .000 (.000) & .018 (.004) & .059 (.006) \\
		\texttt{GPT-3.5-turbo} & .438 (.008) & .158 (.009) & .087 (.006) & .000 (.000) & .180 (.014) & .137 (.005) \\
		\texttt{gemma-3n-E4B} & .519 (.013) & .165 (.010) & .041 (.006) & .082 (.010) & .178 (.015) & .016 (.004) \\
		\texttt{Llama-3.3-70B} & .683 (.011) & .191 (.007) & .014 (.003) & .015 (.003) & .069 (.009) & .030 (.005) \\
		\texttt{Llama-3.1-8B} & .514 (.006) & .174 (.004) & .085 (.004) & .012 (.001) & .103 (.010) & .112 (.003) \\
		\texttt{Llama-4-Maverick} & .673 (.012) & .122 (.007) & .057 (.006) & .005 (.001) & .078 (.009) & .065 (.006) \\
		\texttt{Qwen3-A3B} & .545 (.010) & .216 (.011) & .065 (.008) & .001 (.001) & .052 (.009) & .120 (.007) \\
		\texttt{GPT-4o-mini (reversed)} & .584 (.013) & .227 (.009) & .004 (.001) & .009 (.002) & .120 (.015) & .057 (.006) \\
		\texttt{GPT-4o-mini (temp0)} & .571 (.009) & .246 (.012) & .022 (.004) & .008 (.002) & .097 (.012) & .057 (.005) \\
		\texttt{GPT-4o-mini (multiple extraction)} & .569 (.009) & .218 (.009) & .050 (.004) & .011 (.002) & .095 (.011) & .057 (.005) \\
		\texttt{GPT-4o-mini (in-context)} & .573 (.011) & .271 (.010) & .011 (.003) & .003 (.001) & .088 (.012) & .052 (.006) \\
		\texttt{GPT-4o-mini (assert)} & .559 (.012) & .196 (.013) & .018 (.004) & .027 (.004) & .169 (.019) & .031 (.004) \\
		\texttt{GPT-4o-mini (reflection)} & .586 (.010) & .204 (.012) & .032 (.005) & .008 (.002) & .123 (.013) & .047 (.005) \\
		\texttt{GPT-5.4 (reflection)} & .613 (.009) & .283 (.007) & .002 (.001) & .000 (.000) & .042 (.007) & .060 (.006) \\
		\texttt{GPT-5.4 (third-person reflection)} & .637 (.011) & .240 (.007) & .003 (.001) & .001 (.000) & .063 (.008) & .056 (.006) \\
		\texttt{GPT-5.4 (third-person reflection, medium reasoning)} & .775 (.012) & .119 (.007) & .019 (.003) & .001 (.000) & .008 (.004) & .079 (.007) \\
		\hline
\end{tabular}
}
    \label{tab:sdr summary}
\end{table}

\begin{table}[h!]
    \scriptsize
    \centering
    \caption{\textbf{Mass of polarization with respect to initial stance, for all tested models.} Fraction of the transition mass of the initial stance; mean (SE) over the 112 propositions. A dash marks an initial stance for which the pattern is empty by definition.}
    \resizebox{\textwidth}{!}{% SDR component: Polarization
\begin{tabular}{lccccc}
		\\
		\hline
		Model & Agree strongly & Agree & Neutral & Disagree & Disagree strongly \\
		\hline
		\texttt{GPT-4o-mini} & --- & .589 (.040) & --- & .642 (.037) & --- \\
		\texttt{GPT-4.1} & --- & .397 (.031) & --- & .825 (.028) & --- \\
		\texttt{GPT-5.4} & --- & .777 (.028) & --- & .863 (.023) & --- \\
		\texttt{GPT-3.5-turbo} & --- & .326 (.035) & --- & .462 (.035) & --- \\
		\texttt{gemma-3n-E4B} & --- & .234 (.034) & --- & .592 (.041) & --- \\
		\texttt{Llama-3.3-70B} & --- & .137 (.024) & --- & .817 (.028) & --- \\
		\texttt{Llama-3.1-8B} & --- & .076 (.007) & --- & .795 (.016) & --- \\
		\texttt{Llama-4-Maverick} & --- & .288 (.030) & --- & .322 (.038) & --- \\
		\texttt{Qwen3-A3B} & --- & .526 (.039) & --- & .557 (.042) & --- \\
		\texttt{GPT-4o-mini (reversed)} & --- & .617 (.041) & --- & .519 (.036) & --- \\
		\texttt{GPT-4o-mini (temp0)} & --- & .590 (.040) & --- & .638 (.038) & --- \\
		\texttt{GPT-4o-mini (multiple extraction)} & --- & .498 (.038) & --- & .591 (.026) & --- \\
		\texttt{GPT-4o-mini (in-context)} & --- & .579 (.039) & --- & .777 (.032) & --- \\
		\texttt{GPT-4o-mini (assert)} & --- & .557 (.041) & --- & .423 (.038) & --- \\
		\texttt{GPT-4o-mini (reflection)} & --- & .500 (.041) & --- & .518 (.039) & --- \\
		\texttt{GPT-5.4 (reflection)} & --- & .573 (.038) & --- & .844 (.024) & --- \\
		\texttt{GPT-5.4 (third-person reflection)} & --- & .448 (.038) & --- & .754 (.031) & --- \\
		\texttt{GPT-5.4 (third-person reflection, medium reasoning)} & --- & .267 (.027) & --- & .327 (.032) & --- \\
		\hline
\end{tabular}
}
    \label{tab:sdr polarization}
\end{table}

\begin{table}[h!]
    \scriptsize
    \centering
    \caption{\textbf{Mass of moderation with respect to initial stance, for all tested models.} Fraction of the transition mass of the initial stance; mean (SE) over the 112 propositions. A dash marks an initial stance for which the pattern is empty by definition.}
    \resizebox{\textwidth}{!}{% SDR component: Moderation
\begin{tabular}{lccccc}
		\\
		\hline
		Model & Agree strongly & Agree & Neutral & Disagree & Disagree strongly \\
		\hline
		\texttt{GPT-4o-mini} & .000 (.000) & --- & --- & --- & .108 (.021) \\
		\texttt{GPT-4.1} & .038 (.014) & --- & --- & --- & .081 (.013) \\
		\texttt{GPT-5.4} & .002 (.001) & --- & --- & --- & .008 (.004) \\
		\texttt{GPT-3.5-turbo} & .294 (.027) & --- & --- & --- & .140 (.021) \\
		\texttt{gemma-3n-E4B} & .140 (.023) & --- & --- & --- & .063 (.017) \\
		\texttt{Llama-3.3-70B} & .046 (.015) & --- & --- & --- & .022 (.006) \\
		\texttt{Llama-3.1-8B} & .379 (.020) & --- & --- & --- & .048 (.005) \\
		\texttt{Llama-4-Maverick} & .143 (.024) & --- & --- & --- & .144 (.023) \\
		\texttt{Qwen3-A3B} & .003 (.002) & --- & --- & --- & .322 (.039) \\
		\texttt{GPT-4o-mini (reversed)} & .007 (.004) & --- & --- & --- & .013 (.006) \\
		\texttt{GPT-4o-mini (temp0)} & .000 (.000) & --- & --- & --- & .109 (.021) \\
		\texttt{GPT-4o-mini (multiple extraction)} & .039 (.004) & --- & --- & --- & .209 (.018) \\
		\texttt{GPT-4o-mini (in-context)} & .000 (.000) & --- & --- & --- & .057 (.013) \\
		\texttt{GPT-4o-mini (assert)} & .000 (.000) & --- & --- & --- & .088 (.020) \\
		\texttt{GPT-4o-mini (reflection)} & .000 (.000) & --- & --- & --- & .162 (.027) \\
		\texttt{GPT-5.4 (reflection)} & .002 (.002) & --- & --- & --- & .006 (.003) \\
		\texttt{GPT-5.4 (third-person reflection)} & .002 (.001) & --- & --- & --- & .014 (.006) \\
		\texttt{GPT-5.4 (third-person reflection, medium reasoning)} & .017 (.007) & --- & --- & --- & .076 (.015) \\
		\hline
\end{tabular}
}
    \label{tab:sdr moderation}
\end{table}

\begin{table}[h!]
    \scriptsize
    \centering
    \caption{\textbf{Mass of moderation to neutral with respect to initial stance, for all tested models.} Fraction of the transition mass of the initial stance; mean (SE) over the 112 propositions. A dash marks an initial stance for which the pattern is empty by definition.}
    \resizebox{\textwidth}{!}{% SDR component: Moderation to neutral
\begin{tabular}{lccccc}
		\\
		\hline
		Model & Agree strongly & Agree & Neutral & Disagree & Disagree strongly \\
		\hline
		\texttt{GPT-4o-mini} & .000 (.000) & .039 (.009) & --- & .004 (.002) & .000 (.000) \\
		\texttt{GPT-4.1} & .000 (.000) & .003 (.001) & --- & .002 (.001) & .000 (.000) \\
		\texttt{GPT-5.4} & .000 (.000) & .000 (.000) & --- & .000 (.000) & .000 (.000) \\
		\texttt{GPT-3.5-turbo} & .000 (.000) & .000 (.000) & --- & .001 (.000) & .000 (.000) \\
		\texttt{gemma-3n-E4B} & .027 (.009) & .315 (.035) & --- & .043 (.014) & .023 (.010) \\
		\texttt{Llama-3.3-70B} & .000 (.000) & .073 (.013) & --- & .001 (.001) & .000 (.000) \\
		\texttt{Llama-3.1-8B} & .006 (.001) & .037 (.003) & --- & .011 (.002) & .004 (.001) \\
		\texttt{Llama-4-Maverick} & .000 (.000) & .004 (.001) & --- & .019 (.007) & .001 (.000) \\
		\texttt{Qwen3-A3B} & .000 (.000) & .005 (.004) & --- & .000 (.000) & .000 (.000) \\
		\texttt{GPT-4o-mini (reversed)} & .000 (.000) & .009 (.003) & --- & .034 (.010) & .000 (.000) \\
		\texttt{GPT-4o-mini (temp0)} & .000 (.000) & .038 (.009) & --- & .004 (.002) & .000 (.000) \\
		\texttt{GPT-4o-mini (multiple extraction)} & .000 (.000) & .051 (.009) & --- & .004 (.001) & .000 (.000) \\
		\texttt{GPT-4o-mini (in-context)} & .000 (.000) & .016 (.004) & --- & .001 (.001) & .000 (.000) \\
		\texttt{GPT-4o-mini (assert)} & .000 (.000) & .117 (.018) & --- & .018 (.005) & .001 (.000) \\
		\texttt{GPT-4o-mini (reflection)} & .000 (.000) & .028 (.005) & --- & .010 (.005) & .000 (.000) \\
		\texttt{GPT-5.4 (reflection)} & .000 (.000) & .000 (.000) & --- & .000 (.000) & .000 (.000) \\
		\texttt{GPT-5.4 (third-person reflection)} & .000 (.000) & .003 (.002) & --- & .000 (.000) & .000 (.000) \\
		\texttt{GPT-5.4 (third-person reflection, medium reasoning)} & .003 (.001) & .001 (.000) & --- & .000 (.000) & .002 (.001) \\
		\hline
\end{tabular}
}
    \label{tab:sdr moderation to neutral}
\end{table}

\begin{table}[h!]
    \scriptsize
    \centering
    \caption{\textbf{Mass of flipping with respect to initial stance, for all tested models.} Fraction of the transition mass of the initial stance; mean (SE) over the 112 propositions. A dash marks an initial stance for which the pattern is empty by definition.}
    \resizebox{\textwidth}{!}{% SDR component: Flipping
\begin{tabular}{lccccc}
		\\
		\hline
		Model & Agree strongly & Agree & Neutral & Disagree & Disagree strongly \\
		\hline
		\texttt{GPT-4o-mini} & .067 (.021) & .264 (.034) & --- & .045 (.016) & .108 (.027) \\
		\texttt{GPT-4.1} & .000 (.000) & .002 (.001) & --- & .005 (.002) & .016 (.005) \\
		\texttt{GPT-5.4} & .002 (.001) & .076 (.018) & --- & .001 (.001) & .011 (.009) \\
		\texttt{GPT-3.5-turbo} & .085 (.018) & .190 (.028) & --- & .293 (.031) & .332 (.033) \\
		\texttt{gemma-3n-E4B} & .237 (.035) & .103 (.022) & --- & .251 (.039) & .296 (.040) \\
		\texttt{Llama-3.3-70B} & .000 (.000) & .286 (.038) & --- & .021 (.010) & .036 (.014) \\
		\texttt{Llama-3.1-8B} & .247 (.029) & .119 (.012) & --- & .090 (.011) & .061 (.009) \\
		\texttt{Llama-4-Maverick} & .001 (.000) & .034 (.011) & --- & .018 (.008) & .340 (.042) \\
		\texttt{Qwen3-A3B} & .058 (.020) & .201 (.034) & --- & .002 (.002) & .000 (.000) \\
		\texttt{GPT-4o-mini (reversed)} & .162 (.033) & .321 (.040) & --- & .036 (.013) & .079 (.023) \\
		\texttt{GPT-4o-mini (temp0)} & .067 (.021) & .264 (.034) & --- & .045 (.016) & .108 (.027) \\
		\texttt{GPT-4o-mini (multiple extraction)} & .058 (.019) & .255 (.033) & --- & .050 (.016) & .111 (.026) \\
		\texttt{GPT-4o-mini (in-context)} & .078 (.023) & .313 (.038) & --- & .022 (.010) & .029 (.014) \\
		\texttt{GPT-4o-mini (assert)} & .075 (.020) & .090 (.018) & --- & .324 (.039) & .358 (.041) \\
		\texttt{GPT-4o-mini (reflection)} & .100 (.026) & .357 (.037) & --- & .039 (.014) & .121 (.028) \\
		\texttt{GPT-5.4 (reflection)} & .007 (.005) & .192 (.032) & --- & .000 (.000) & .009 (.009) \\
		\texttt{GPT-5.4 (third-person reflection)} & .019 (.010) & .252 (.035) & --- & .002 (.001) & .040 (.016) \\
		\texttt{GPT-5.4 (third-person reflection, medium reasoning)} & .019 (.011) & .010 (.006) & --- & .007 (.003) & .004 (.004) \\
		\hline
\end{tabular}
}
    \label{tab:sdr flipping}
\end{table}

\begin{table}[h!]
    \scriptsize
    \centering
    \caption{\textbf{Mass of deviation from neutrality with respect to initial stance, for all tested models.} Fraction of the transition mass of the initial stance; mean (SE) over the 112 propositions. A dash marks an initial stance for which the pattern is empty by definition.}
    \resizebox{\textwidth}{!}{% SDR component: Deviation from neutrality
\begin{tabular}{lccccc}
		\\
		\hline
		Model & Agree strongly & Agree & Neutral & Disagree & Disagree strongly \\
		\hline
		\texttt{GPT-4o-mini} & --- & --- & .283 (.026) & --- & --- \\
		\texttt{GPT-4.1} & --- & --- & .200 (.026) & --- & --- \\
		\texttt{GPT-5.4} & --- & --- & .297 (.032) & --- & --- \\
		\texttt{GPT-3.5-turbo} & --- & --- & .687 (.023) & --- & --- \\
		\texttt{gemma-3n-E4B} & --- & --- & .080 (.019) & --- & --- \\
		\texttt{Llama-3.3-70B} & --- & --- & .149 (.023) & --- & --- \\
		\texttt{Llama-3.1-8B} & --- & --- & .558 (.014) & --- & --- \\
		\texttt{Llama-4-Maverick} & --- & --- & .325 (.032) & --- & --- \\
		\texttt{Qwen3-A3B} & --- & --- & .598 (.034) & --- & --- \\
		\texttt{GPT-4o-mini (reversed)} & --- & --- & .283 (.028) & --- & --- \\
		\texttt{GPT-4o-mini (temp0)} & --- & --- & .284 (.026) & --- & --- \\
		\texttt{GPT-4o-mini (multiple extraction)} & --- & --- & .287 (.026) & --- & --- \\
		\texttt{GPT-4o-mini (in-context)} & --- & --- & .261 (.030) & --- & --- \\
		\texttt{GPT-4o-mini (assert)} & --- & --- & .155 (.020) & --- & --- \\
		\texttt{GPT-4o-mini (reflection)} & --- & --- & .233 (.024) & --- & --- \\
		\texttt{GPT-5.4 (reflection)} & --- & --- & .301 (.032) & --- & --- \\
		\texttt{GPT-5.4 (third-person reflection)} & --- & --- & .279 (.031) & --- & --- \\
		\texttt{GPT-5.4 (third-person reflection, medium reasoning)} & --- & --- & .393 (.033) & --- & --- \\
		\hline
\end{tabular}
}
    \label{tab:sdr deviation}
\end{table}

\begin{figure}[h!]
    \centering
    \begin{tabular}[b]{cc}
        \small (\textbf{A}) &
        \includegraphics[width=.85\textwidth]{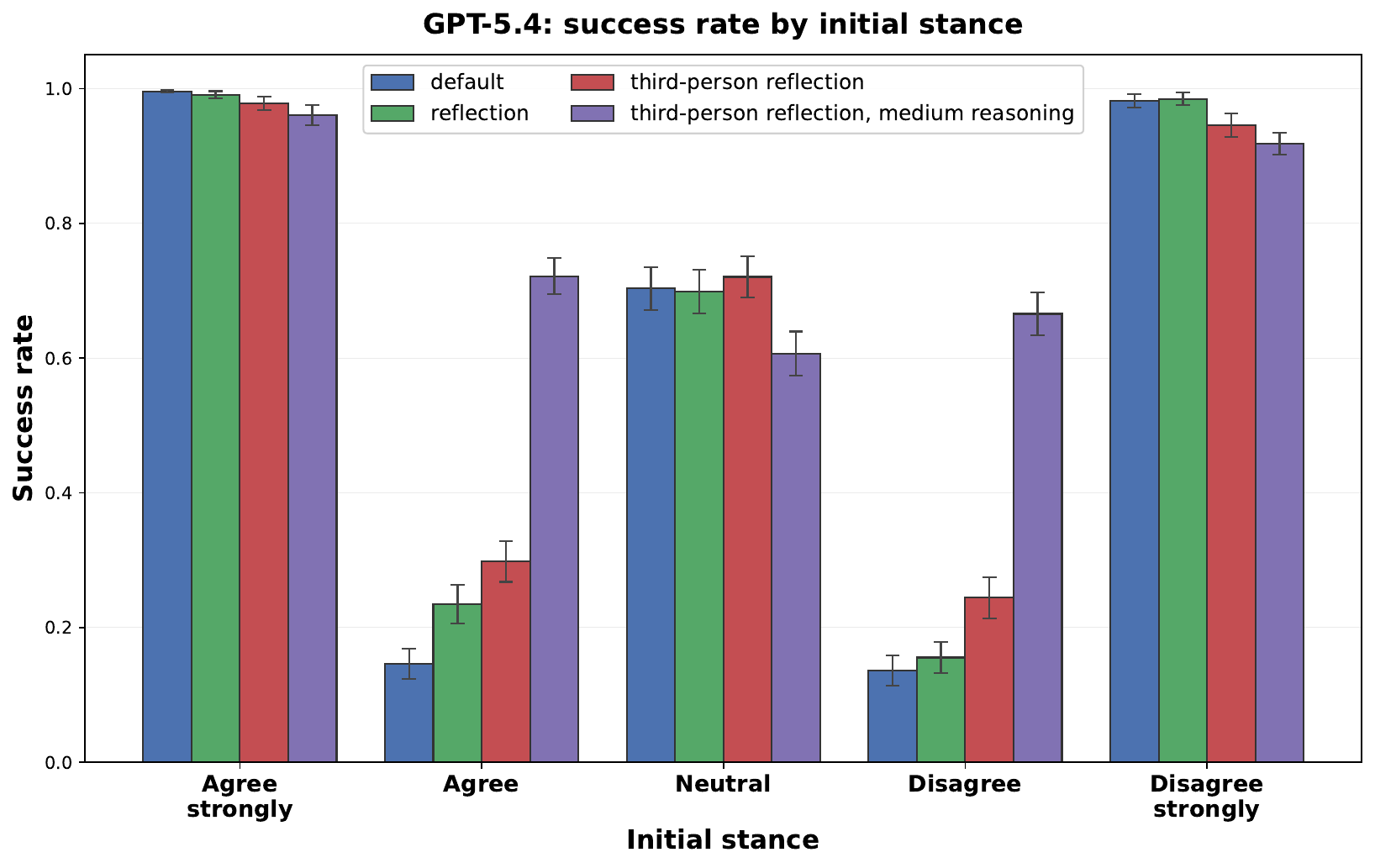}
    \end{tabular}\\
    \vspace{.8em}
    \begin{tabular}[b]{cc}
      \small (\textbf{B}) &
      \includegraphics[width=.85\textwidth]{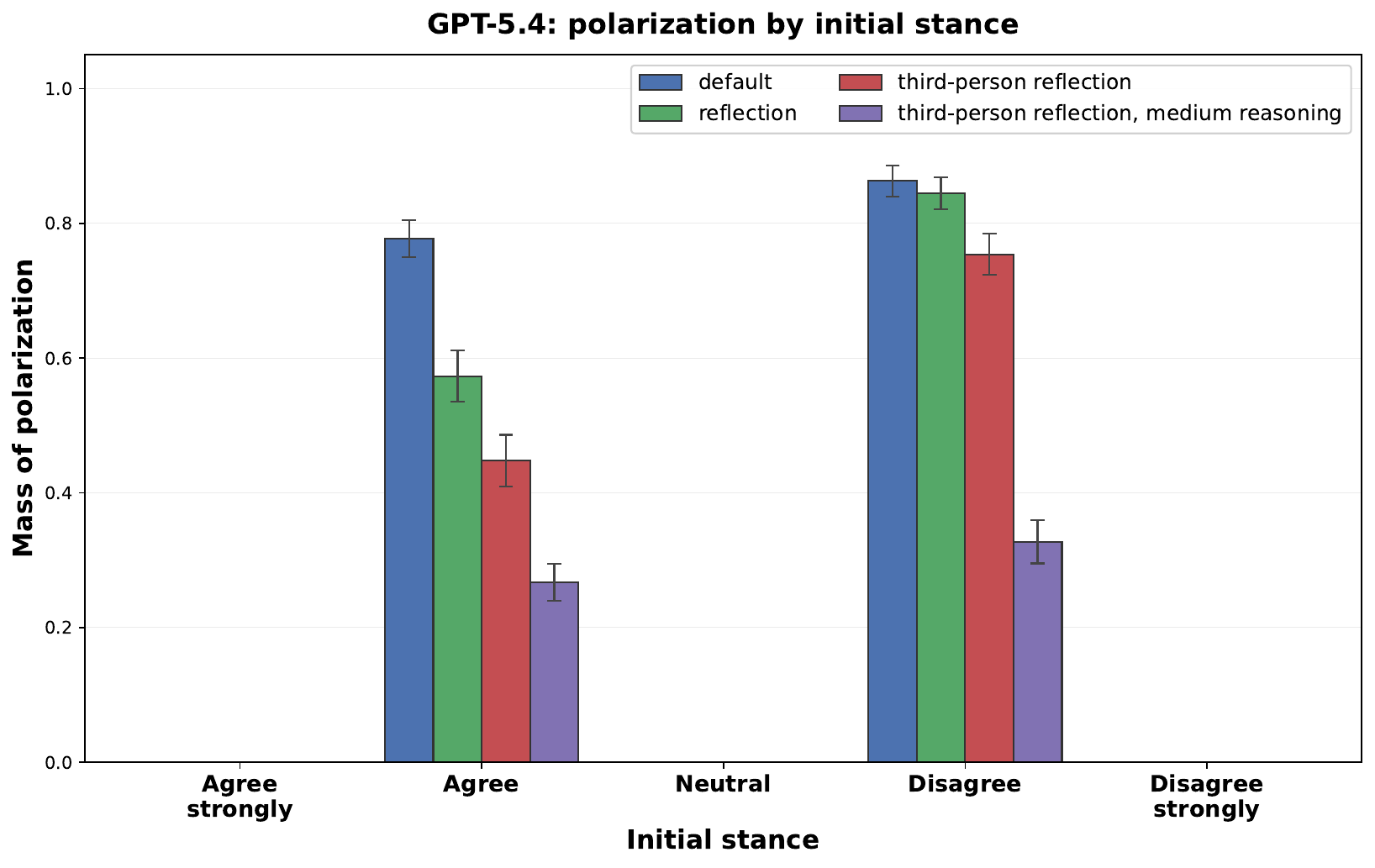}
    \end{tabular}
    \caption{
        \textbf{\texttt{GPT-5.4} across the four configurations, with respect to initial stance.} (\textbf{A}) Success rate. (\textbf{B}) Mass of polarization. The configurations are the default configuration, reflection, third-person reflection, and third-person reflection with medium reasoning effort; the three later configurations each add one change to the previous one. Bars are means over the 112 propositions and error bars are standard errors. The values are listed in Tables~\ref{tab:success rates} and \ref{tab:sdr polarization}.
    }
    \label{fig:gpt54 configurations}
\end{figure}

\begin{table}[h!]
    \scriptsize
    \centering
    \caption{\textbf{Argument counts by drift stage and initial stance for the human-extraction study.} Each cell records the number of arguments (out of 20 per initial stance) attributable to each drift stage, for proposition ``The use of AI should be abandoned'' generated by \texttt{GPT-5.4}. Generation-stage drift: the human-majority label differs from the known initial stance, indicating that the generated argument itself does not carry the intended stance signal. Extraction-stage drift: the AI-majority label differs from the human-majority label, attributing the discrepancy to the LLM extractor rather than the generated content. The two rows are not mutually exclusive.}
    % Human-extraction stage counts (SI Table S11): generated by analysis/make_drift_examples.py; do not edit.
\begin{tabular}{lccccc}
		\\
		\hline
		Initial Stance & Agree strongly & Agree & Neutral & Disagree & Disagree strongly \\
		\hline
		Generation-stage drift & 0 & 20 & 5 & 0 & 1 \\
		Extraction-stage drift & 0 & 19 & 13 & 20 & 1 \\
		\hline
\end{tabular}

    \label{tab:human extraction}
\end{table}

\begin{table}[h!]
    \scriptsize
    \centering
    \caption{\textbf{One example argument per drift pattern from the human-extraction study.} All examples come from the single proposition of the study, ``The use of AI should be abandoned'', with arguments generated by \texttt{GPT-5.4} and AI labels extracted by \texttt{GPT-5.4}; they illustrate the patterns on this proposition and do not estimate their frequency. Drift pattern: the pattern of the transition from the initial stance to the AI-majority label, as defined in the main text. Drift stage: generation stage if the human-majority label differs from the initial stance, extraction stage if the AI-majority label differs from the human-majority label, both stages if both differ. For each pattern the argument with the largest sum of the human-majority and AI-majority vote shares is shown, in full; \DriftExampleTiesFlipWord{} arguments share the largest sum for flipping and \DriftExampleTiesDevWord{} for deviation from neutrality, and the first in the row order of the study's data file is shown. Moderation and moderation to neutral do not occur among the 100 arguments of this proposition.}
    \setlength{\tabcolsep}{3pt}
    \resizebox{\textwidth}{!}{% Drift-pattern examples (TASK 3.2): generated by analysis/make_drift_examples.py; do not edit.
% rows = argument index in debata_speeches_3234_filtered_with_human_9encode.xlsx: polarization -> row 78 (rep 18), flipping -> row 25 (rep 5), deviation from neutrality -> row 47 (rep 7)
\begin{tabular}{llllll}
		\hline
		Initial stance & Generated argument & \parbox[t]{1.9cm}{Human-majority label} & \parbox[t]{1.9cm}{AI-majority label} & \parbox[t]{2.6cm}{Drift pattern} & \parbox[t]{3.0cm}{Drift stage} \\
		\hline
		Disagree & \parbox[t]{5.0cm}{Abandoning AI entirely would throw away major benefits in medicine, education, accessibility, scientific research, and public services. Like other powerful technologies, AI can be misused, but that argues for strong rules, transparency, and accountability---not total rejection. The United States would also harm its competitiveness and innovation capacity by stepping back while other countries continue developing it. A better approach is to limit dangerous applications, require human oversight where stakes are high, and enforce ethical standards so society can capture AI's advantages while reducing risks.\vspace{3pt}} & \parbox[t]{1.9cm}{Disagree} & \parbox[t]{1.9cm}{Disagree strongly} & \parbox[t]{2.6cm}{polarization} & \parbox[t]{3.0cm}{extraction stage} \\
		Agree & \parbox[t]{5.0cm}{AI should not be abandoned, but its use should be approached carefully. In the United States, AI already helps doctors detect diseases earlier, assists businesses with routine tasks, and supports scientific research. Rejecting it entirely would mean giving up major benefits and global competitiveness. The real issue is not whether AI should exist, but how to govern it responsibly. Strong rules on privacy, transparency, bias, and accountability can reduce harm while preserving innovation. With thoughtful oversight, AI can remain a valuable tool rather than something society must completely discard.\vspace{3pt}} & \parbox[t]{1.9cm}{Disagree} & \parbox[t]{1.9cm}{Disagree strongly} & \parbox[t]{2.6cm}{flipping} & \parbox[t]{3.0cm}{both stages (flipping at generation, polarization at extraction)} \\
		Neutral & \parbox[t]{5.0cm}{AI's future depends on how it is designed, governed, and used. It can improve medical diagnosis, expand access to education, and automate dangerous tasks, yet it also raises serious concerns about bias, privacy, job disruption, and misuse. Because both benefits and harms are significant, abandoning it entirely may be premature, but embracing it without safeguards would be reckless. A balanced approach would focus on strict oversight, transparency, accountability, and limits on high-risk applications while society continues evaluating whether its advantages truly outweigh its costs.\vspace{3pt}} & \parbox[t]{1.9cm}{Neutral} & \parbox[t]{1.9cm}{Disagree} & \parbox[t]{2.6cm}{deviation from neutrality} & \parbox[t]{3.0cm}{extraction stage} \\
		\hline
\end{tabular}
}
    \label{tab:drift examples}
\end{table}

\begin{table}[p]
    \scriptsize
    \centering
    \rotatebox{90}{\begin{minipage}{0.95\textheight}
    \caption{\textbf{Model identifiers, run dates and extraction distributions of every run.} One row per run of Tables~\ref{tab:success rates}--\ref{tab:pass rate}, in the same order. Model identifier: the model string requested through the API. Run dates: first and last date of the run, in UTC$-$4. Extraction distribution: log probabilities, each trial contributes the normalized probabilities of the five option letters; observed frequencies, each trial contributes the letter found in the extraction output. \texttt{GPT-4o-mini (temp0)} is a re-analysis of the \texttt{GPT-4o-mini} run with each trial's extraction vector replaced by a point mass on its most probable option. The same model performs generation and extraction in every run, over the 112 propositions with 100 trials per initial stance.}
    \centering
    \resizebox{\linewidth}{!}{% Reproducibility table (TASK 3.1): generated by analysis/make_repro_table.py; do not edit.
\begin{tabular}{llllll}
		\hline
		Run & Model identifier & API & Prompt & Run dates (2026, UTC$-$4) & Extraction distribution \\
		\hline
		\parbox[t]{5.2cm}{\texttt{GPT-4o-mini}} & \texttt{gpt-4o-mini} & OpenAI & default & Jan 22--23 & \parbox[t]{3.2cm}{log probabilities} \\
		\parbox[t]{5.2cm}{\texttt{GPT-4.1}} & \texttt{gpt-4.1} & OpenAI & default & Jan 21--22 & \parbox[t]{3.2cm}{log probabilities} \\
		\parbox[t]{5.2cm}{\texttt{GPT-5.4}} & \texttt{gpt-5.4} & OpenAI & default & Mar 10 & \parbox[t]{3.2cm}{observed frequencies} \\
		\parbox[t]{5.2cm}{\texttt{GPT-3.5-turbo}} & \texttt{gpt-3.5-turbo} & OpenAI & default & Jan 23--24 & \parbox[t]{3.2cm}{log probabilities} \\
		\parbox[t]{5.2cm}{\texttt{gemma-3n-E4B}} & \texttt{google/gemma-3n-E4B-it} & Together & default & Jan 25--Feb 3 & \parbox[t]{3.2cm}{log probabilities} \\
		\parbox[t]{5.2cm}{\texttt{Llama-3.3-70B}} & \texttt{meta-llama/Llama-3.3-70B-Instruct-Turbo} & Together & default & Feb 4--11 & \parbox[t]{3.2cm}{observed frequencies} \\
		\parbox[t]{5.2cm}{\texttt{Llama-3.1-8B}} & \texttt{meta-llama/Meta-Llama-3.1-8B-Instruct-Turbo} & Together & default & Feb 4--11 & \parbox[t]{3.2cm}{observed frequencies} \\
		\parbox[t]{5.2cm}{\texttt{Llama-4-Maverick}} & \texttt{meta-llama/Llama-4-Maverick-17B-128E-Instruct-FP8} & Together & default & Feb 3--12 & \parbox[t]{3.2cm}{log probabilities} \\
		\parbox[t]{5.2cm}{\texttt{Qwen3-A3B}} & \texttt{Qwen/Qwen3-Next-80B-A3B-Instruct} & Together & default & Feb 5 & \parbox[t]{3.2cm}{log probabilities} \\
		\hline
		\parbox[t]{5.2cm}{\texttt{GPT-4o-mini (reversed)}} & \texttt{gpt-4o-mini} & OpenAI & default & Jan 24 & \parbox[t]{3.2cm}{log probabilities} \\
		\parbox[t]{5.2cm}{\texttt{GPT-4o-mini (temp0)}} & \texttt{gpt-4o-mini} & OpenAI & default & Jan 22--23 & \parbox[t]{3.2cm}{argmax of the log probabilities} \\
		\parbox[t]{5.2cm}{\texttt{GPT-4o-mini (multiple extraction)}} & \texttt{gpt-4o-mini} & OpenAI & default & Jan 23 & \parbox[t]{3.2cm}{log probabilities} \\
		\parbox[t]{5.2cm}{\texttt{GPT-4o-mini (in-context)}} & \texttt{gpt-4o-mini} & OpenAI & in-context & Jan 23 & \parbox[t]{3.2cm}{log probabilities} \\
		\parbox[t]{5.2cm}{\texttt{GPT-4o-mini (assert)}} & \texttt{gpt-4o-mini} & OpenAI & assertion & Mar 6--14 & \parbox[t]{3.2cm}{log probabilities} \\
		\parbox[t]{5.2cm}{\texttt{GPT-4o-mini (reflection)}} & \texttt{gpt-4o-mini} & OpenAI & reflection & May 24 & \parbox[t]{3.2cm}{log probabilities} \\
		\hline
		\parbox[t]{5.2cm}{\texttt{GPT-5.4 (reflection)}} & \texttt{gpt-5.4} & OpenAI & reflection & May 24 & \parbox[t]{3.2cm}{observed frequencies} \\
		\parbox[t]{5.2cm}{\texttt{GPT-5.4 (third-person reflection)}} & \texttt{gpt-5.4} & OpenAI & third-person reflection & Jun 1 & \parbox[t]{3.2cm}{observed frequencies} \\
		\parbox[t]{5.2cm}{\texttt{GPT-5.4 (third-person reflection, medium reasoning)}} & \texttt{gpt-5.4} & OpenAI & third-person reflection & Jun 2--9 & \parbox[t]{3.2cm}{observed frequencies} \\
		\hline
\end{tabular}
}
    \label{tab:reproducibility}
    \end{minipage}}
\end{table}

% The tables of this section are placed before its text (layout only, TASK 4.1).
\clearpage
\subsection{Positional bias}
As pointed out in previous studies, the order in which answer choices are presented can significantly influence respondents' selections, a phenomenon known as positional bias~\citep{pezeshkpour2024order,zheng2024mcq}. However, it is hard to remove positional bias due to identifiability issues and it is not practical to iterate over all possible permutations of answer choices. Besides randomly shuffling the order of answer choices in the multiple extraction strategy we implemented, we could take another perspective, following the most existing surveys and questionnaires and ordering answer choices from positive to negative (e.g., Agree strongly, Agree, Neutral, Disagree, Disagree strongly)~\citep{johns2010likert}, treating positional bias of such order as effect from the communication practices in the real world.  

For robustness, we also conduct experiments with the reversed order of answer choices to examine the impact of positional bias on LLMs' responses. Although we could observe the effect of positional bias by comparing the results in Table~\ref{tab:success rates}, we find that the SPR of $\SPRGPTfourominiReversed$ (Fig.~\ref{fig:SPR_all_models}) remains relatively stable compared to the original order setting ($\SPRGPTfouromini$).  

\subsection{Temperature}
One might argue that the failure of recovering the initial stance is due to the randomness introduced by the temperature setting during extraction. A higher temperature allows for more diverse outputs, which could lead to higher chances of deviating from the initial stance. To address this concern, we conduct additional analysis of our raw data results by adjusting the probability of the most probable answer to 1, and the rest to 0, which is equivalent to setting the temperature to 0 during extraction. Yet the SPR remains at $\SPRGPTfourominiTempZero$ (Fig.~\ref{fig:SPR_all_models}), indicating that the randomness introduced by temperature is not a significant factor in the stance preservation performance of LLMs.

\subsection{Pass rate and AUC}\label{sec:pass rate}

We define the pass rate at a threshold: a proposition--initial stance combination passes the threshold if its success rate, the diagonal probability averaged over its trials, exceeds the threshold, and the pass rate is the proportion of combinations that pass. Table~\ref{tab:pass rate} presents the pass rate at threshold $0.95$ by initial stance, as the proportion of the 112 propositions that pass, offering a more relaxed criterion than perfect preservation.  % lint:ok

\begin{figure}[ht]
    \centering
    \begin{tabular}[b]{cc}
        \small (\textbf{A}) &
        \includegraphics[width=.9\textwidth]{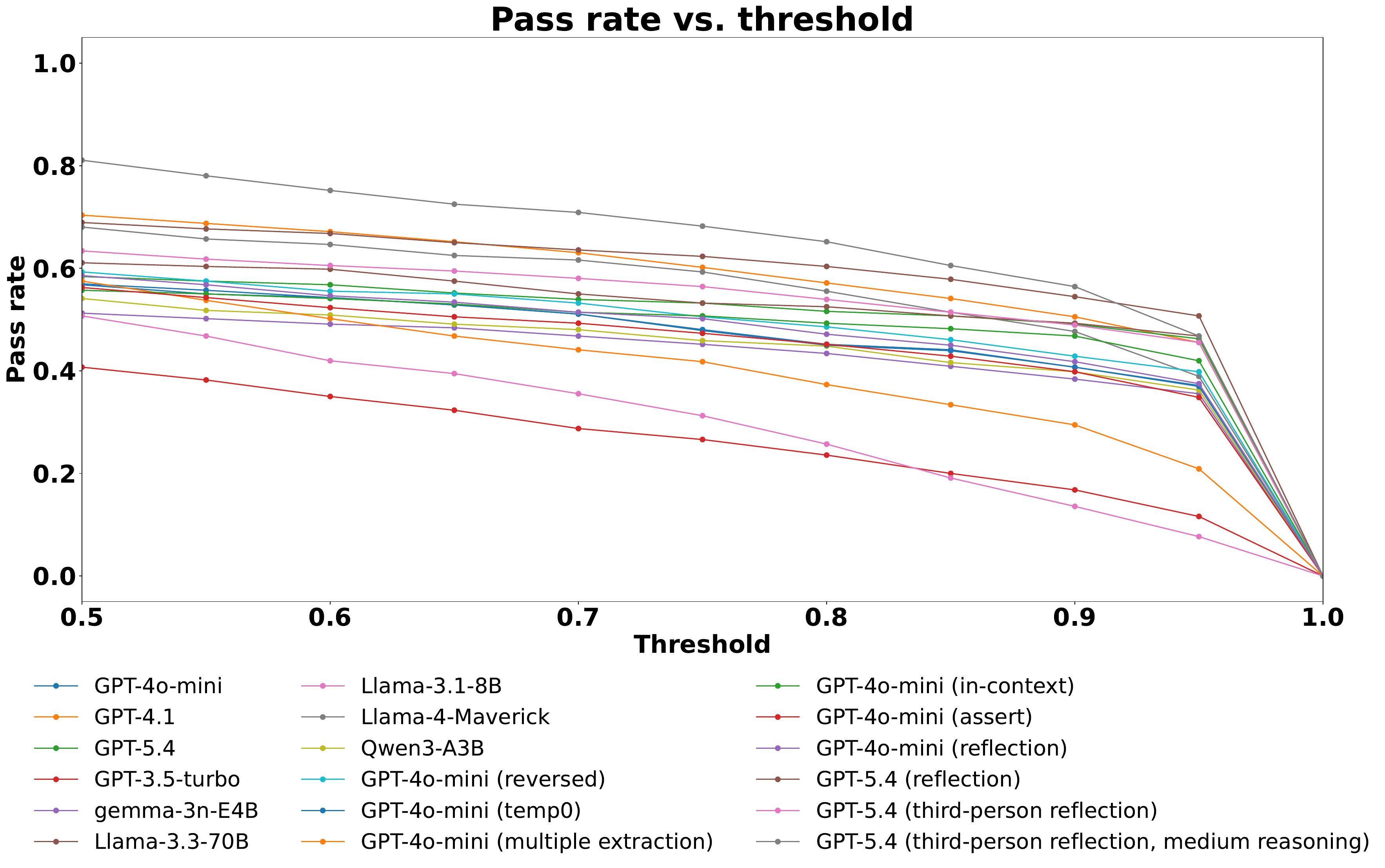}
    \end{tabular}\\
    \vspace{.8em}
    \begin{tabular}[b]{cc}
      \small (\textbf{B})&
      \includegraphics[width=0.9\textwidth]{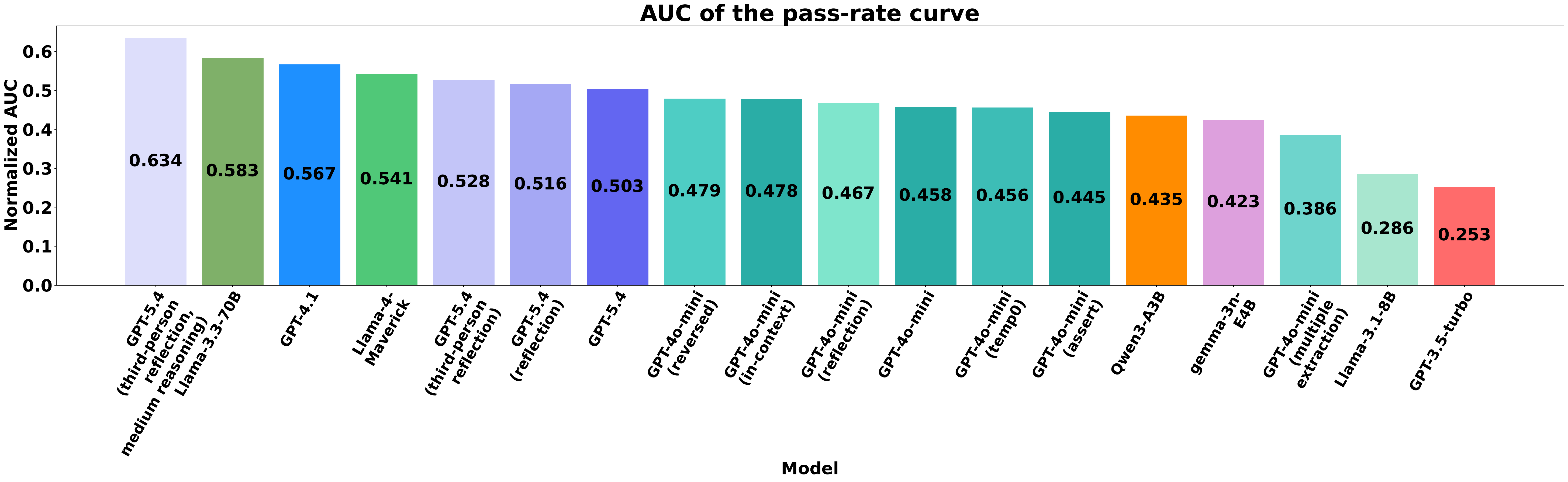} 
    \end{tabular}
    \caption{
        \textbf{Pass rate and AUC.} (\textbf{A}) Pass rate as a function of the threshold from $0.5$ to $1$, over all propositions and initial stances. (\textbf{B}) Area under the pass-rate curve of (A), normalized by its maximum value of $0.5$, across different models, providing a summary measure of stance preservation performance over the range of thresholds. % lint:ok
    }
  \label{fig:AUC}
\end{figure}

We further compute the area under the curve (AUC) of the pass rate as the threshold varies from $0.5$ to $1$ in increments of $0.05$, over all propositions and initial stances (Fig.~\ref{fig:AUC}A), normalized by its maximum value of $0.5$ so that it lies between $0$ and $1$. This AUC serves as a comprehensive summary statistic (Fig.~\ref{fig:AUC}B), capturing the model's ability to preserve initial stances across a continuum of strictness levels, analogous to our SPR metric. % lint:ok

\end{document}